\documentclass[10pt,twocolumn,letterpaper]{article}

\usepackage{iccv}              

\usepackage[dvipsnames]{xcolor}

\definecolor{darkgreen}{rgb}{0, 0.4, 0}
\definecolor{darkred}{rgb}{0.5, 0.1, 0.1}
\definecolor{darkblu}{rgb}{0, 0, 0.6}
\definecolor{bgblu}{rgb}{0.9, 0.96, 1}
\definecolor{pergreen}{rgb}{0, 0.7, 0}

\definecolor{barblue}{HTML}{1f77b4}
\definecolor{bargreen}{HTML}{2ca02c}
\definecolor{barorange}{HTML}{ff7f0e}

\definecolor{iccvblue}{rgb}{0.21,0.49,0.74}
\usepackage[pagebackref,breaklinks,colorlinks,citecolor=iccvblue]{hyperref}
\usepackage{amsmath}
\usepackage{multirow}
\usepackage{multicol}

\usepackage[ruled,vlined]{algorithm2e}
\usepackage{array,multirow,graphicx}
\usepackage{xr}
\usepackage[skip=0pt]{caption}
\usepackage{floatrow}

\usepackage[export]{adjustbox}

\newcommand{\ours}{\texttt{ANSWER}\@\xspace}

\newcommand{\cfg}{\texttt{CFG}\@\xspace}
\newcommand{\dns}{\texttt{DNS}\@\xspace}
\newcommand{\cnp}{\texttt{CNP}\@\xspace}
\newcommand{\np}{\texttt{NP}\@\xspace}
\newcommand{\dnp}{\texttt{DNP}\@\xspace}

\definecolor{iccvblue}{rgb}{0.21,0.49,0.74}

\def\paperID{8378} 
\def\confName{ICCV}
\def\confYear{2025}

\title{Guiding Diffusion Models with Adaptive Negative Sampling\\ Without External Resources}

\author{Alakh Desai \quad Nuno Vasconcelos\\
University of California, San Diego\\
{\tt\small \{ahdesai, nvasconcelos\}@ucsd.edu}
}

\begin{document}
\twocolumn[{
\maketitle
\vspace{-.2in}
\centering
\resizebox{\linewidth}{!}{
\begin{tabular}{c@{\hskip 0.05em}c@{\hskip 0.1em}c@{\hskip 0.05em}c@{\hskip 0.1em}c@{\hskip 0.05em}c@{\hskip 0.1em}c@{\hskip 0.05em}c}

\includegraphics[width=.18\linewidth,valign=m]{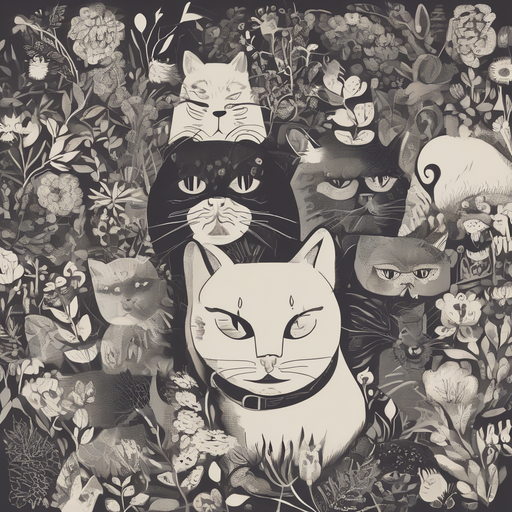} & 
\includegraphics[width=.18\linewidth,valign=m]{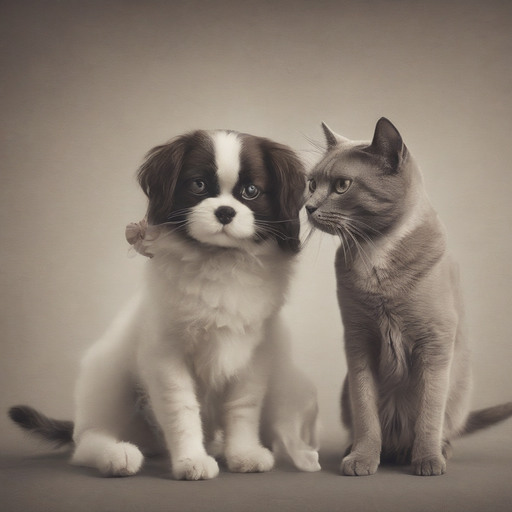} & 
\includegraphics[width=.18\linewidth,valign=m]{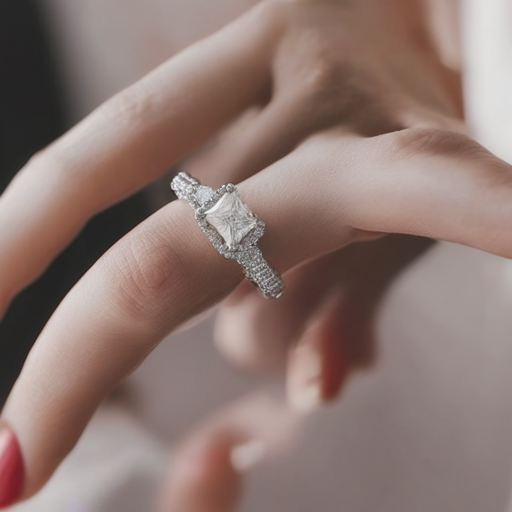} & 
\includegraphics[width=.18\linewidth,valign=m]{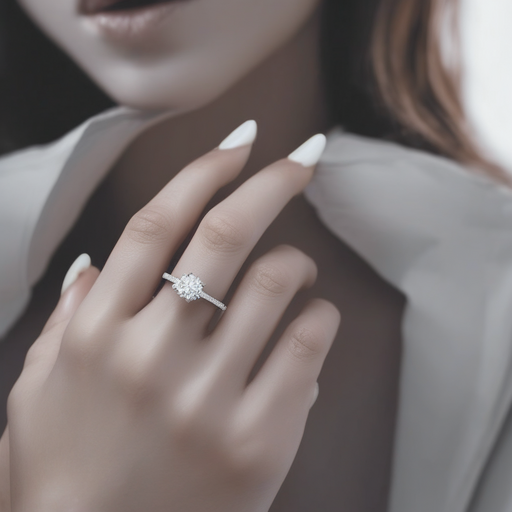} &
\includegraphics[width=.18\linewidth,valign=m]{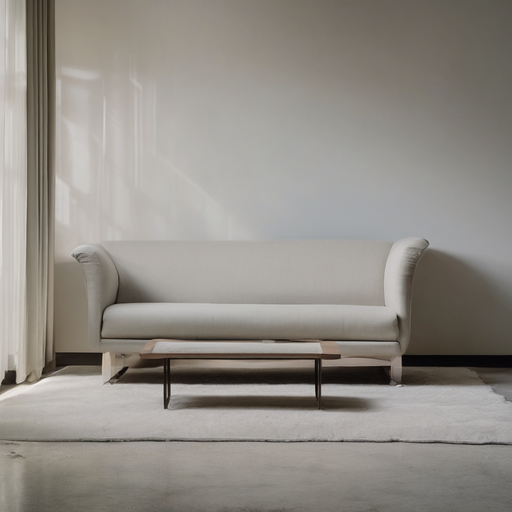} & 
\includegraphics[width=.18\linewidth,valign=m]{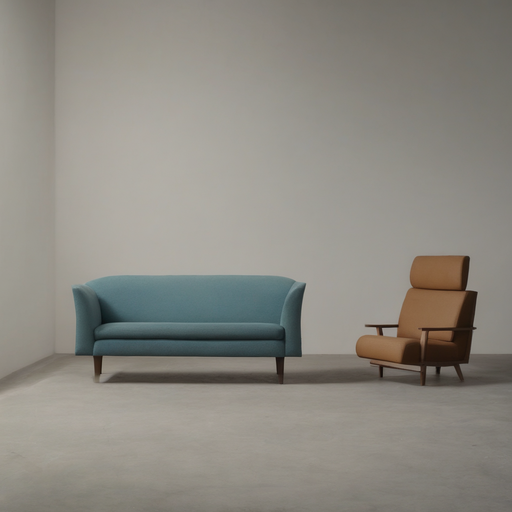} &
\includegraphics[width=.18\linewidth,valign=m]{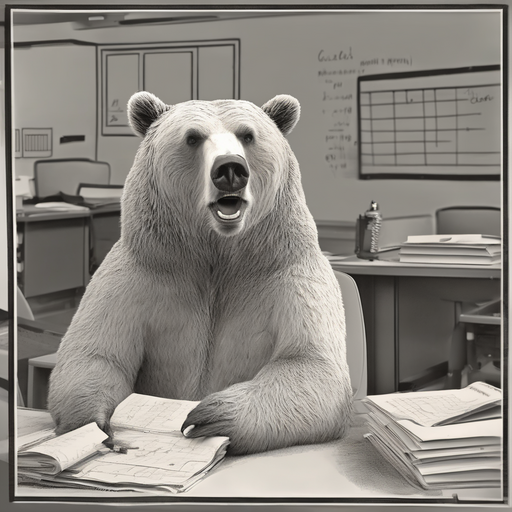} & 
\includegraphics[width=.18\linewidth,valign=m]{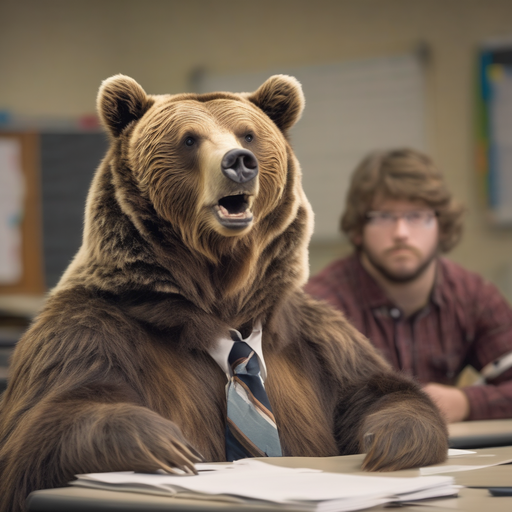} \\

\multicolumn{2}{c}{a cat and a dog} & \multicolumn{2}{c}{a diamond ring on a girl's hand} & \multicolumn{2}{c}{ a couch on the left of a chair} & \multicolumn{2}{c}{a confused grizzly bear in calculus class}\\[0.03in]

\includegraphics[width=.18\linewidth,valign=m]{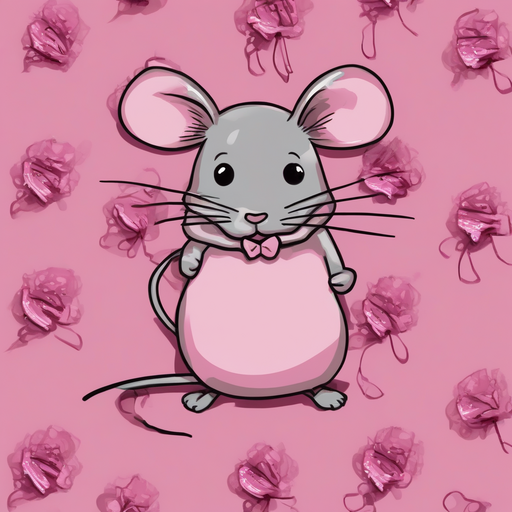} &
\includegraphics[width=.18\linewidth,valign=m]{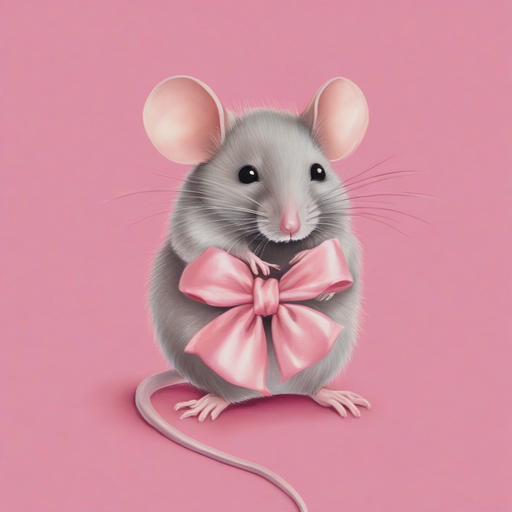} & 
\includegraphics[width=.18\linewidth,valign=m]{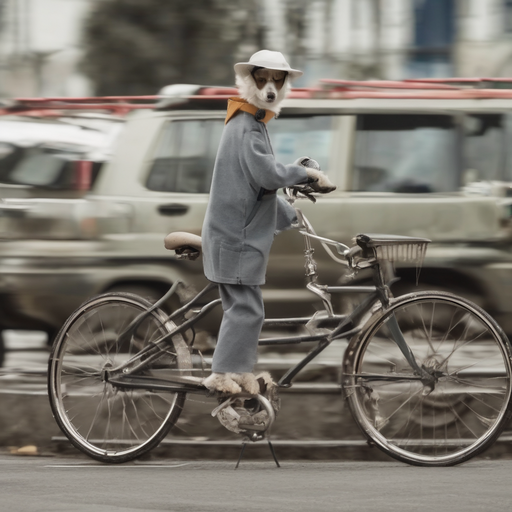} &
\includegraphics[width=.18\linewidth,valign=m]{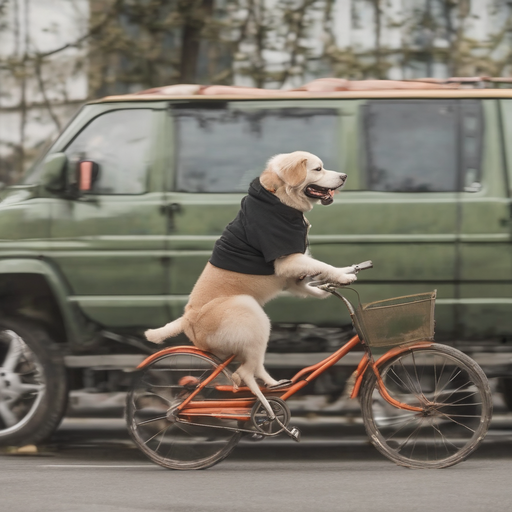} &
\includegraphics[width=.18\linewidth,valign=m]{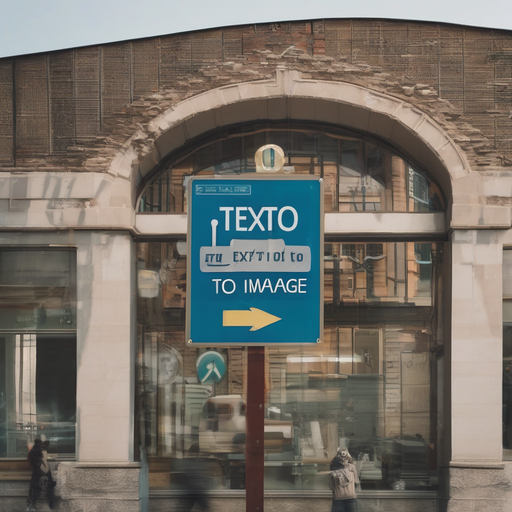} &
\includegraphics[width=.18\linewidth,valign=m]{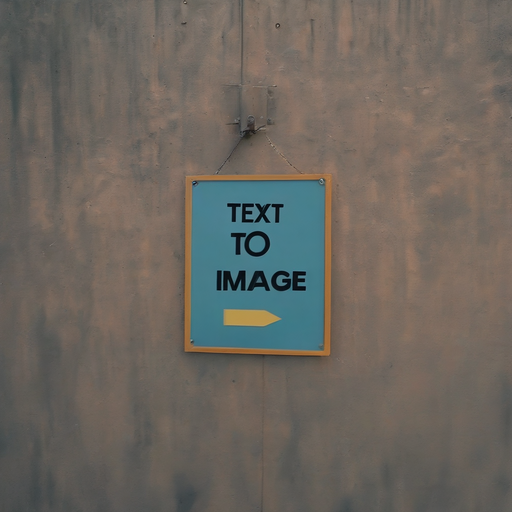} &
\includegraphics[width=.18\linewidth,valign=m]{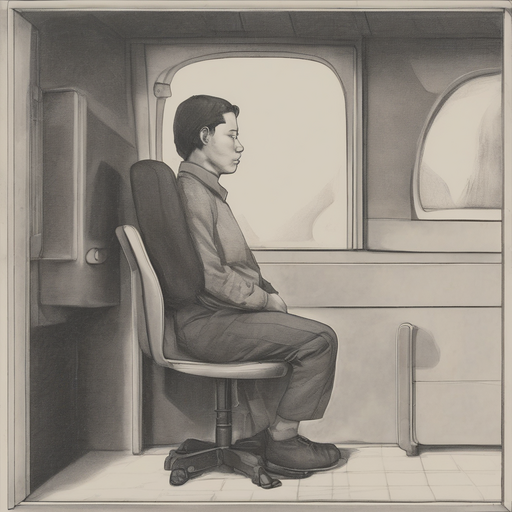} & 
\includegraphics[width=.18\linewidth,valign=m]{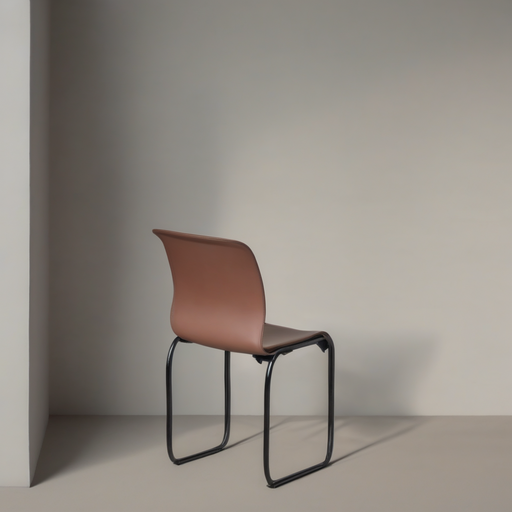}\\

\multicolumn{2}{c}{a mouse and a pink bow} & \multicolumn{2}{c}{a dog riding bicycle} & \multicolumn{2}{c}{a sign that says `Text to Image'} & \multicolumn{2}{c}{seat for a person with a back and four legs}\\[0.03in]

\includegraphics[width=.18\linewidth,valign=m]{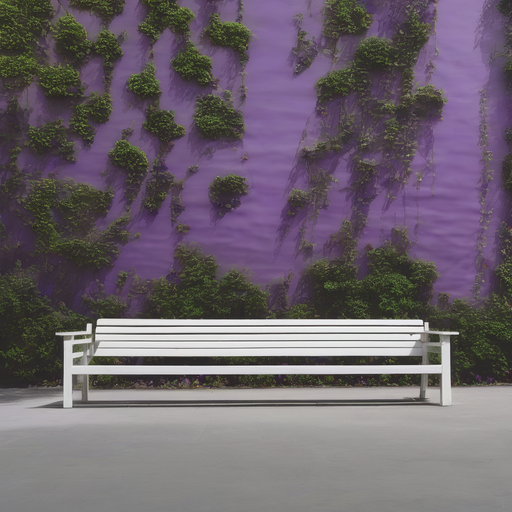} &
\includegraphics[width=.18\linewidth,valign=m]{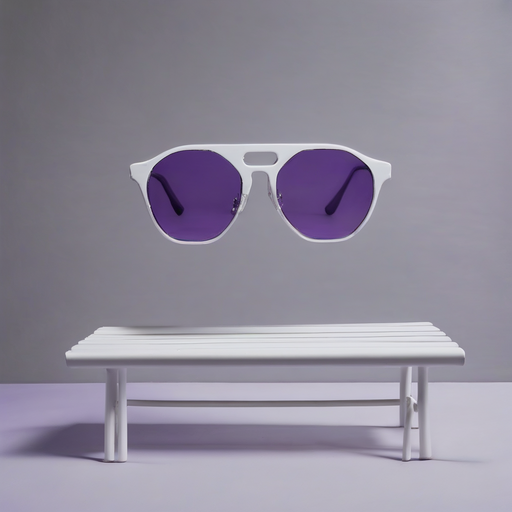} & 
\includegraphics[width=.18\linewidth,valign=m]{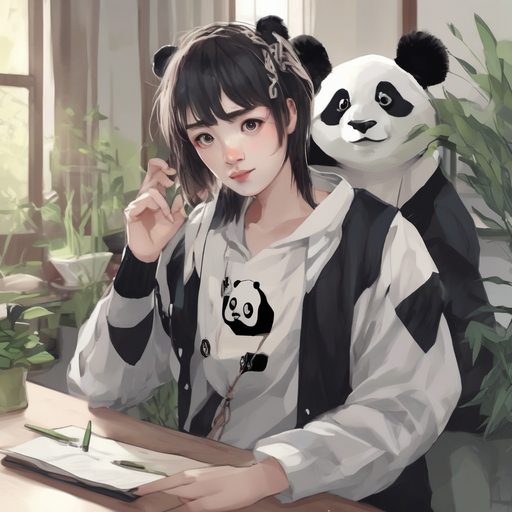} &
\includegraphics[width=.18\linewidth,valign=m]{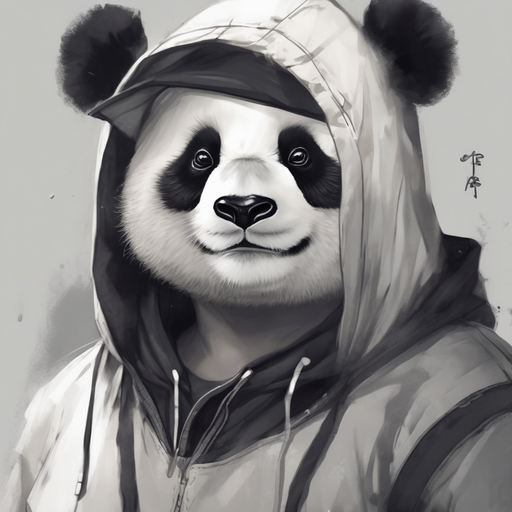} &
\includegraphics[width=.18\linewidth,valign=m]{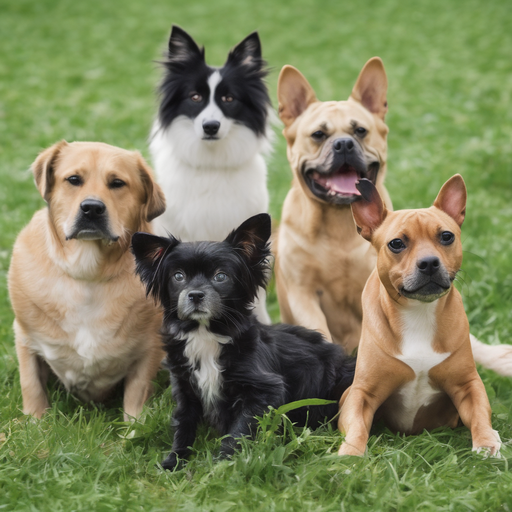} &
\includegraphics[width=.18\linewidth,valign=m]{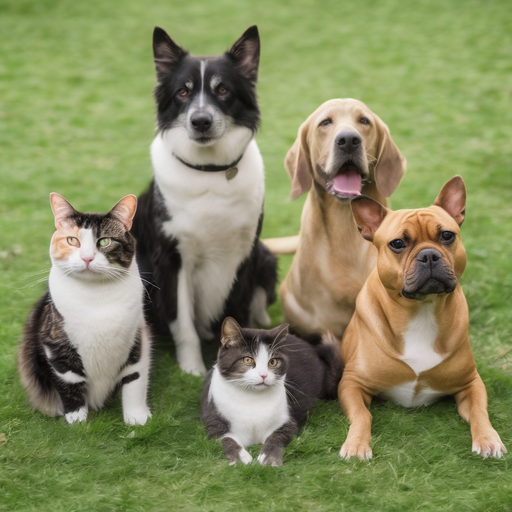} &
\includegraphics[width=.18\linewidth,valign=m]{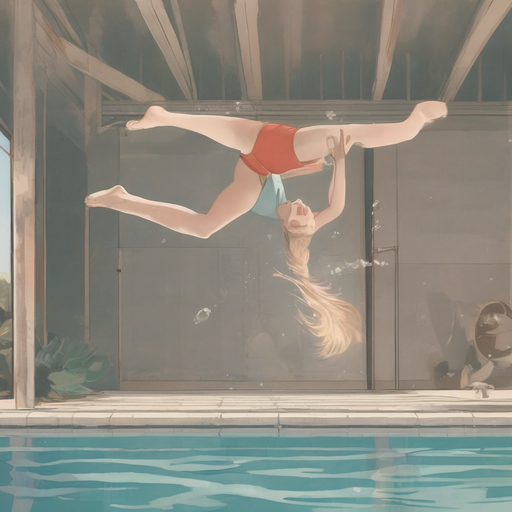} & 
\includegraphics[width=.18\linewidth,valign=m]{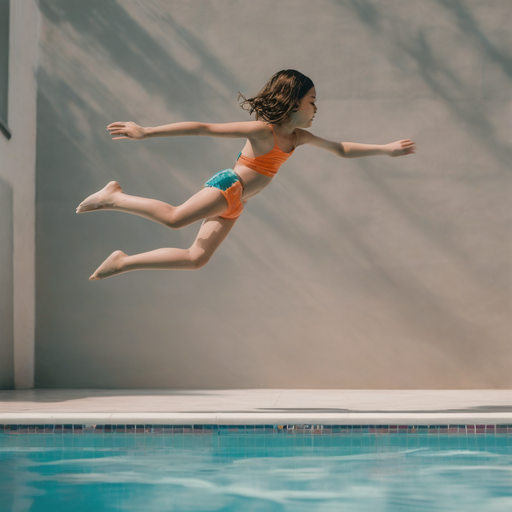}\\

\multicolumn{2}{c}{a white bench and purple glasses} & \multicolumn{2}{c}{a panda as a human} & \multicolumn{2}{c}{two cats and three dogs sitting on the grass} & \multicolumn{2}{c}{a girl diving into the pool}\\[0.03in]

\end{tabular}}
\captionof{figure}{
    Images synthesized by SDXL~(\cfg) (Left) vs. SDXL+\ours (Right) for the prompt displayed below each image pair. \ours optimizes the negative guidance at each diffusion step to maximize image quality and prompt compliance.
    }
\label{fig:teaser}
\vspace{.1in}
}]
\begin{abstract}
Diffusion models (DMs) have demonstrated an unparalleled ability to create diverse and high-fidelity images from text prompts. However, they are also well-known to vary substantially regarding both prompt adherence and quality. Negative prompting was introduced to improve prompt compliance by specifying what an image must not contain. Previous works have shown the existence of an ideal negative prompt that can maximize the odds of the positive prompt. In this work, we explore relations between negative prompting and classifier-free guidance (\cfg) to develop a sampling procedure, {\it Adaptive Negative Sampling Without External Resources\/} (\ours), that accounts for both positive and negative conditions from a single prompt. This leverages the internal understanding of negation by the diffusion model to increase the odds of generating images faithful to the prompt. \ours is a training-free technique, applicable to any model that supports \cfg, and allows for negative grounding of image concepts without an explicit negative prompts, which are lossy and incomplete. Experiments show that adding \ours to existing DMs outperforms the baselines on multiple benchmarks and is preferred by humans $2 \times$ more over the other methods. 

\end{abstract}
\vspace{-0.2in}
\section{Introduction}
\label{sec:intro}

\begin{figure}[t]
\centering
\resizebox{\linewidth}{!}{
\begin{tabular}{c@{\hskip 0.03em}c@{\hskip 0.03em}c@{\hskip 0.03em}c@{\hskip 0.03em}c}
\cfg$@5$ & \cfg$@10$ & \cfg$@20$  & $DNP$ & $ANSWER$\\
\includegraphics[width=0.25\linewidth]{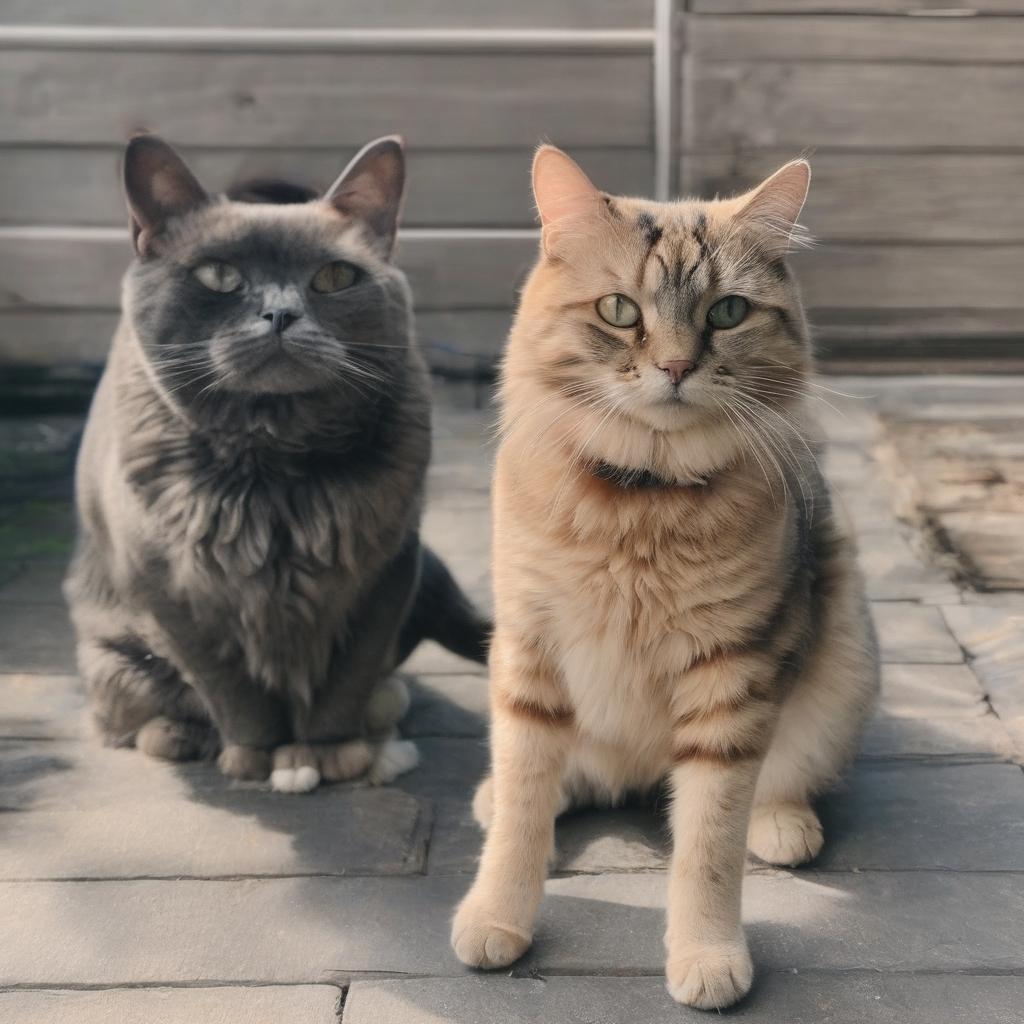}
&\includegraphics[width=0.25\linewidth]{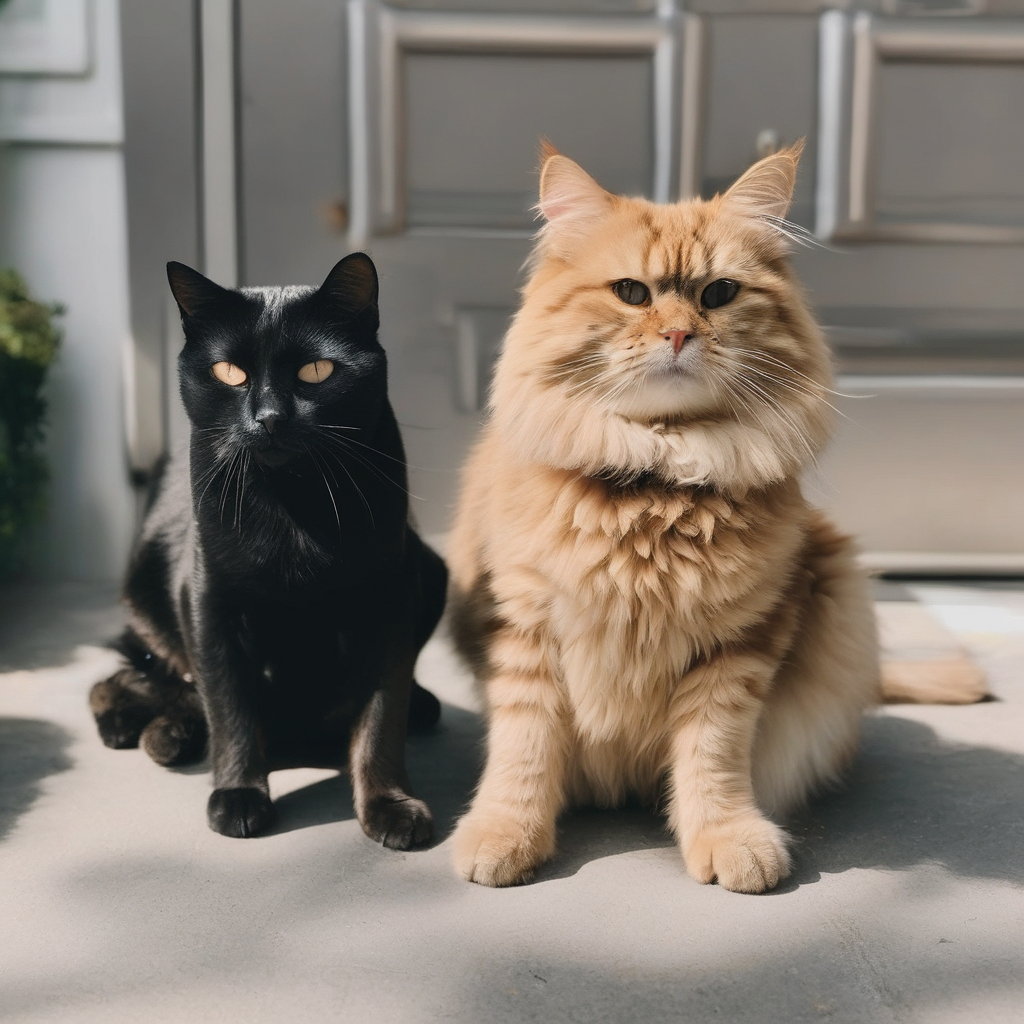}
&\includegraphics[width=0.25\linewidth]{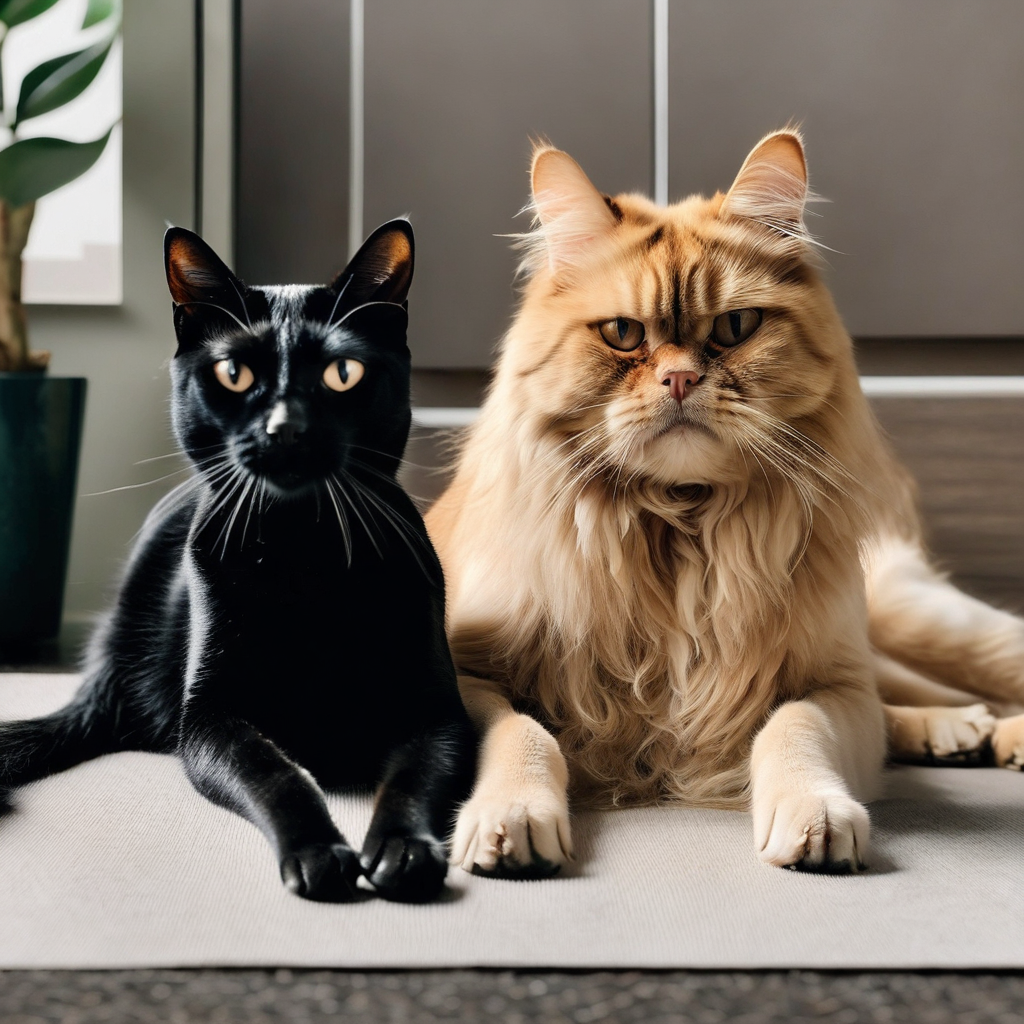}
&\includegraphics[width=0.25\linewidth]{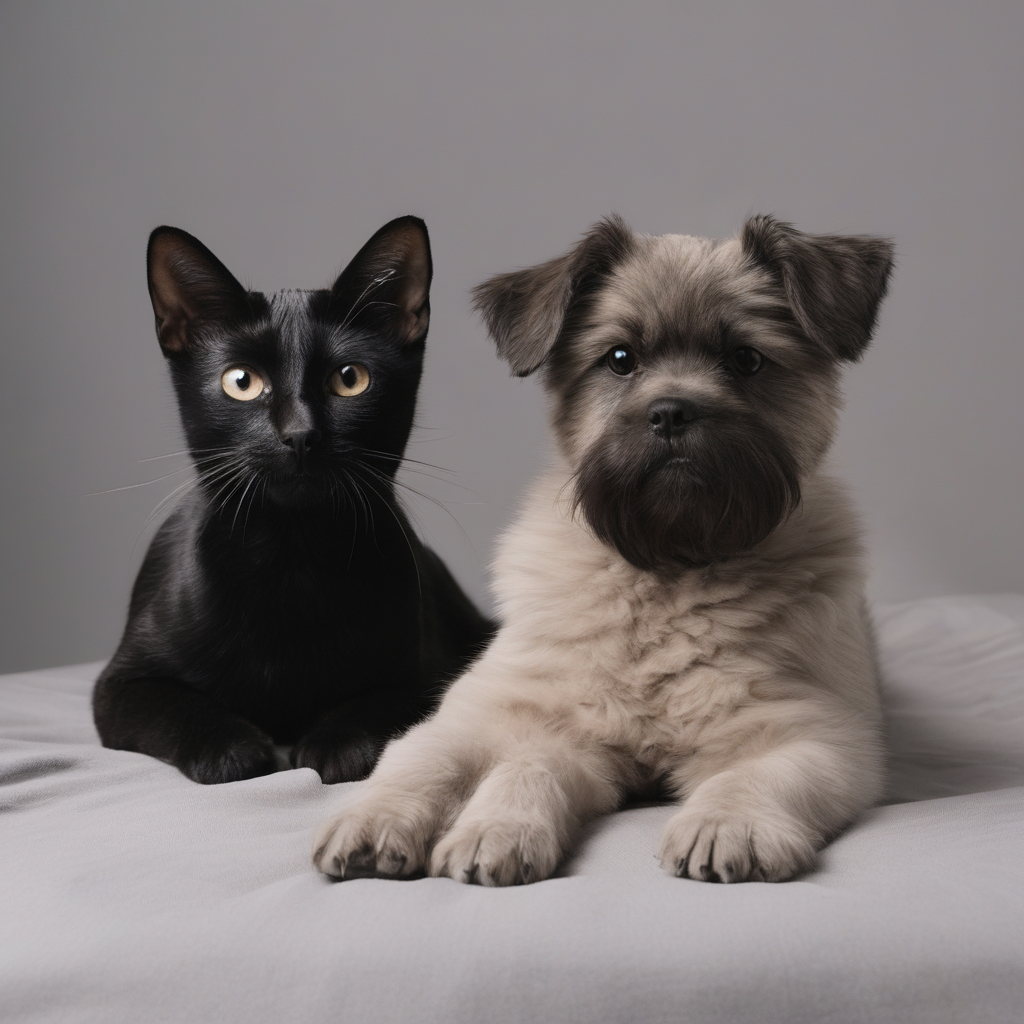}
&\includegraphics[width=0.25\linewidth]{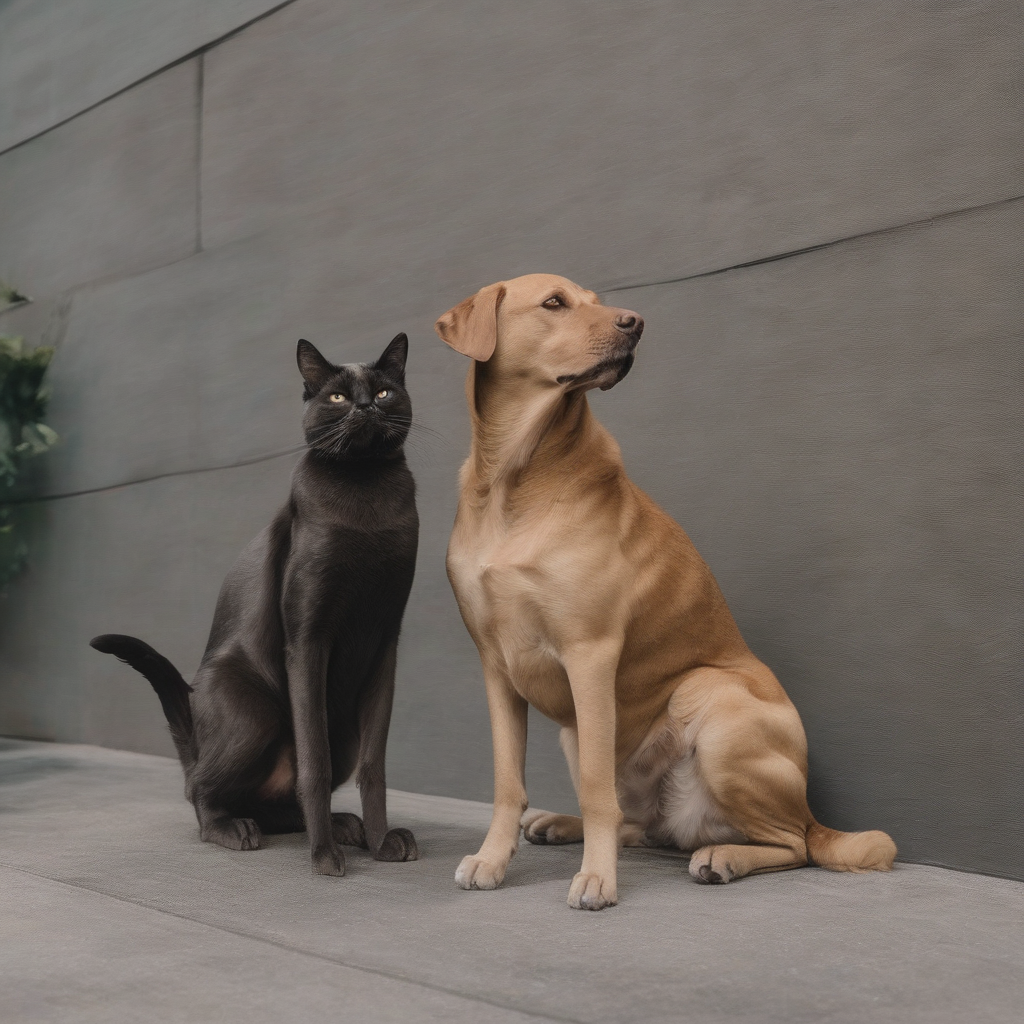}
\\[-0.07cm]
\end{tabular}}
\vspace{-0.15in}
\caption{Effect of guidance scale and different sampling methods on the prompt {\bf p} = {\tt a cat on the left of a dog}. }\label{fig:samples}
\end{figure}

Diffusion models (DMs), such as DALL-E 2~\cite{dalle2}, Imagen~\cite{imagen}, or SDXL~\cite{sdxl}, have shown a remarkable ability to generate compelling images from textual prompts, and are now considered state-of-the-art image synthesis. However, while the quality of the synthesized images is usually excellent, these methods still suffer from limited control over the image semantics. As shown in \Cref{fig:samples}, the generated images frequently fail to comply with the prompt (e.g. ``two cats'' instead of ``a cat on the left of a dog''). This has motivated a literature on DMs that condition image synthesis on user-specified visual constraints, such as bounding boxes~\cite{gligen, groundeddiff, layoutdiff}, sketches~\cite{controlnet, sketch}, etc. While effective, these approaches have limitations of their own. First, it is difficult to devise a universal visual conditioning module. While bounding boxes can easily specify object locations, it is much less clear what visual user input can help a model produce a scene that ``exudes happiness'' or ``is artsy". Second, while visual specifications are suitable for professional image creators, they are too cumbersome for casual users. 

An alternative is to modify the sampling process of the DM. A popular sampling technique is classifier-free guidance (\cfg)~\cite{cfg}. This combines samples from a conditioned and an unconditioned network to improve the trade-off between image diversity and prompt compliance, under the control of a guidance scale parameter $s$. While increasing $s$ tends to improve prompt adherence, it is usually insufficient to guarantee compliance with complex prompts, e.g, involving spatial relations between multiple objects. In the example of \Cref{fig:samples}, even a high guidance scale ($s=20$) creates an image of ``two cats". This problem motivated various techniques that optimize the DM latent during the denoising process~\cite{ae,syngen} to maximize prompt compliance. However, this alters the Markov chain implemented by the model, creating a mismatch between training and inference that can easily result in low-quality images. 

A less drastic manipulation of the sampling process is to use negative prompting~\cite{compos}. This allows the user to specify an additional prompt of what the image {\it is not\/} about. Sampling techniques like classifier-free guidance (\cfg)~\cite{cfg} are then modified to use positive and negative text prompts to improve prompt compliance. While negative prompting is a popular and successful mechanism for removing specific concepts, it is frequently difficult to specify negative prompts (e.g., the negative of the prompt of \Cref{fig:samples} is unclear). This has motivated efforts to determine the negative prompt automatically. \dnp~\cite{dnp} has shown that this can be done by sampling a negative image using the DM itself, which is then captioned to produce a negative prompt. This was shown to outperform latent optimization methods and can be quite successful, as illustrated in \Cref{fig:samples}, where \dnp generates an image better aligned with the prompt (apart from the dog's extra leg) than \cfg, even for a small guidance scale ($s=6$). The authors of \cite{dnp} observe a semantic mismatch between successful negative prompts generated in this way and the human definition of a negative. In \Cref{fig:samples}, the negative prompt used to create the \dnp image is ``a colorful pattern with flowers and leaves". This confirms the difficulty of generating good negative prompts.  

While \dnp is effective, it requires running the denoising process twice and captioning the negative image with an external visual-language model. This has three major limitations: (1) the added complexity of the external model, (2) the discrete approximation of the negative image by a text prompt that fails to capture many of its details, and (3) the fact that negation is only estimated once (first run of the DM), rather than iteratively throughout the diffusion process of the negatively prompted model (second run). We will show that the negation prompt can change significantly as the diffusion chain progresses. 

We propose a new {\it Adaptive Negative Sampling Without External Resources\/} (\ours) procedure, which addresses all these limitations without explicit user input or model retraining. We hypothesize that the capacity of negative prompting to alter the DM's Markov chain can be utilized in more powerful ways by continuously adapting the negative hypothesis during sampling, thus accounting for the changing nature of the image being denoised. Instead of producing a single negative prompt that conditions all denoising steps, we {\it dynamically apply negative guidance at each diffusion step} to estimate the negative noise that {\it matches} the noisy image synthesized at that step. As a result, the negation is performed  {\it adaptively throughout the chain\/}. This makes \ours an enhanced DM sampling mechanism that, as illustrated in \Cref{fig:samples}, tends to generate images that better comply with the prompt than \cfg and \dnp. There is also no need for translation into a language prompt, eliminating the associated loss of information and the need for external captioning. 

Overall, this work makes three main contributions:
\begin{itemize}
     \item It is the first to investigate DM sampling schemes that involve a negative hypothesis produced adaptively to match the denoised image at each diffusion step. It is fully implemented in the DM latent space, thus avoiding costly and incomplete translation to text. 
     \item We introduce the fully autonomous \ours sampling procedure, which requires no negative input by the DM user and no model retraining.
    \item Through extensive experiments on various datasets, we demonstrate that \ours can be seamlessly integrated into existing DMs, offering a practical solution to reduce the semantic gap between humans and DMs.
\end{itemize}

\begin{figure*}[t]
\centering 
\includegraphics[width=\linewidth]{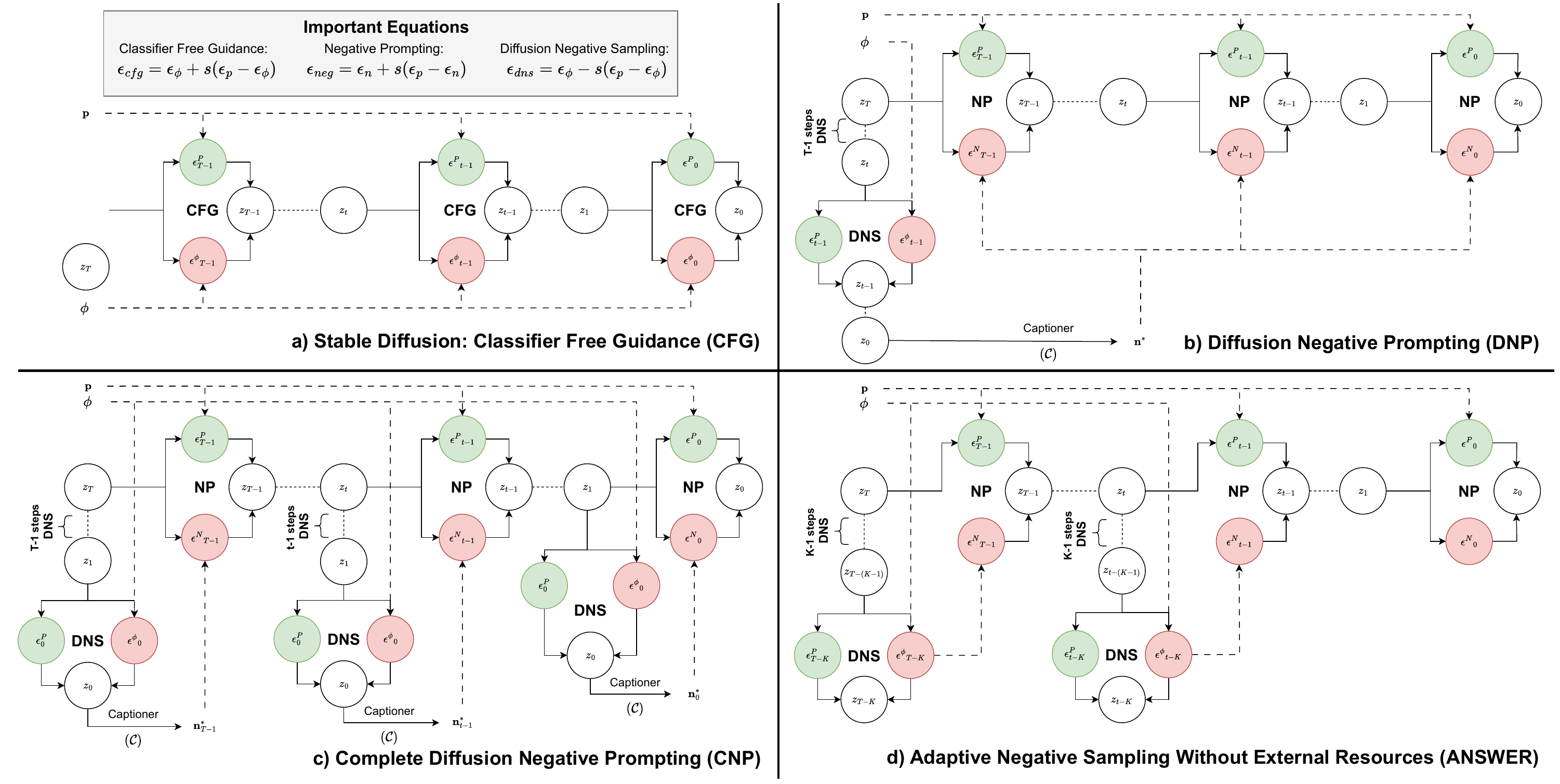}
\vspace{-0.3in}
\caption{Sampling approaches discussed in this work. \dnp, \cnp, and \ours all rely on a \dns chain to generate the negative condition. While \dnp, and \cnp use an external captioning model $\cal C$ to produce negative text prompts, \ours performs the negative conditioning in the latent space of the DM. \cnp runs a complete \dns chain at each diffusion iteration $t$, whereas \ours only requires $K$ \dns iterations.}
\label{fig:overview}
\end{figure*}

\section{Preliminaries}\label{sec:Preliminaries}

\subsection{Diffusion Models}
A DM is a denoising model that synthesizes an image by sequentially denoising a noise code or {\it seed}. DM training involves a combination of a forward and a backward process. The forward process is a Markov chain that maps an image into a seed by gradually adding noise. The backward process performs step-by-step denoising, using a neural network $\epsilon_\theta$ that predicts the noise added at each forward step. Latent DMs are a popular class of DM that uses an encoder/decoder pair $(\mathcal{E}(.), \mathcal{D}(.))$ to produce a low-dimensional latent space $\cal Z$, where diffusion takes place. The forward and backward processes are as follows.

\noindent\textbf{Forward  Process:} A noisy latent representation is obtained at each time step $t$, with $z_t = \sqrt{\bar{\alpha}_t} z + \sqrt{1-\bar{\alpha}_t} \epsilon$, where $\epsilon \sim \mathcal{N}(0, I)$. Here, $\bar{\alpha}_t = \prod_{s=1}^{t}(1-\beta_s)$, with $\{\beta_1,...,\beta_T\}$ fixed according to a variance schedule. During training, the network $\epsilon_\theta$ learns to predict the noise added ($\epsilon_t$) to the latent representation $z$ at each time step $t$. This prediction is conditioned by the embedding $\tau({\bf p})$ of a text prompt $\bf p$, where $\tau(.)$ is a text encoder. The following loss is then minimized to learn the parameters $\theta$ of the denoising network
\begin{equation}
    \mathcal{L} = \mathbb{E}_{t\in[1,T],\epsilon_t\sim\mathcal{N}(0,I)}\Bigl[  \lVert \epsilon_t - \epsilon_\theta(z_t;t,\tau({\bf p})) \rVert^2 \Bigr],
\end{equation}
where $\mathcal{N}$ represents the Gaussian distribution.

\noindent\textbf{Backward  Process:} Image generation is performed by iteratively alternating between denoising and sampling with the learned network $\epsilon_\theta$
\begin{equation}
    \hat{\epsilon}_p = \epsilon_\theta(z_t;t,\tau({\bf p}))
        \label{eq:hatep}
\end{equation}
A sampling method \cite{ddpm, ddim, euler} is then used to obtain $z_{t-1}$ from $z_t$ and $\hat{\epsilon}_p$. A noise seed initializes $z_T \sim \mathcal{N}(0, I)$ to produce a latent $z_0$. The final image is obtained by passing the latent to the decoder, $x_0 = \mathcal{D}(z_0)$. 

\noindent\textbf{Energy-based model interpretation:} Langevin Dynamics \cite{song2019generative, langevindyn} enable an energy-based interpretation of DM sampling. If $E_\theta$ is an energy function, sampling from DM can be viewed as sampling from the probability distribution, $p_\theta(z_t|{\bf p}) \propto e^{-E_\theta(z_t;t,\tau({\bf p}))}$ and training the DM to learn the noise $\epsilon_t$, can be viewed as learning the negative of the score function, $\nabla_{z_t} \log p_\theta(z_t|\tau({\bf p}))=-\epsilon_\theta(z_t;t,\tau({\bf p}))/\sqrt{1-\bar{\alpha}_t}$


\subsection{Guidance}
Using \cref{eq:hatep} in the DM backward process can produce images of weak compliance with prompt $\bf p$. Several enhancements have been proposed. 

\begin{figure*}[t]
\centering
\resizebox{\linewidth}{!}{
\begin{tabular}{c@{\hskip 0.05em}c@{\hskip 0.05em}c@{\hskip 0.05em}c@{\hskip 0.05em}c@{\hskip 0.05em}c@{\hskip 0.05em}c@{\hskip 0.05em}c@{\hskip 0.05em}c@{\hskip 0.05em}c}
$t=0$ & $t=2$ & $t=3$ & $t=4$ & $t=5$ & $t=7$ & $t=12$ & $t=15$ & $t=35$ & $t=40$ \\
\includegraphics[width=0.15\linewidth]{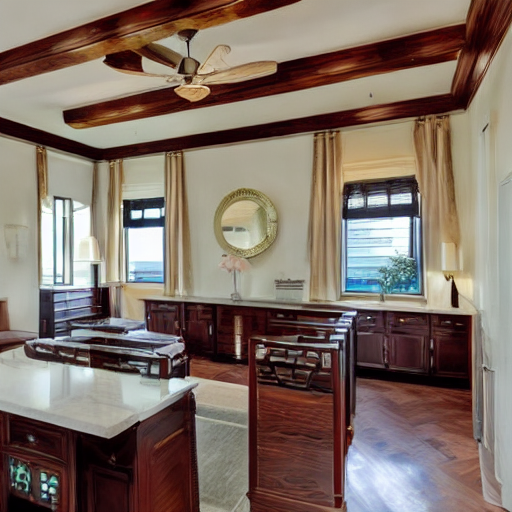}
&\includegraphics[width=0.15\linewidth]{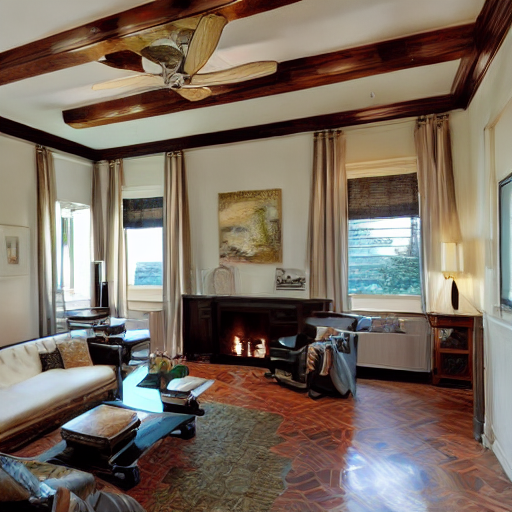}
&\includegraphics[width=0.15\linewidth]{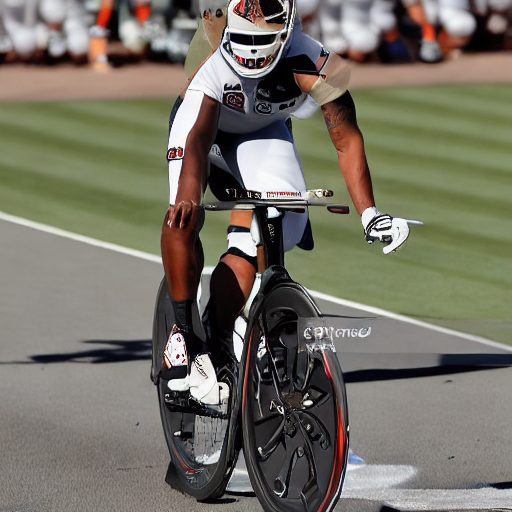}
&\includegraphics[width=0.15\linewidth]{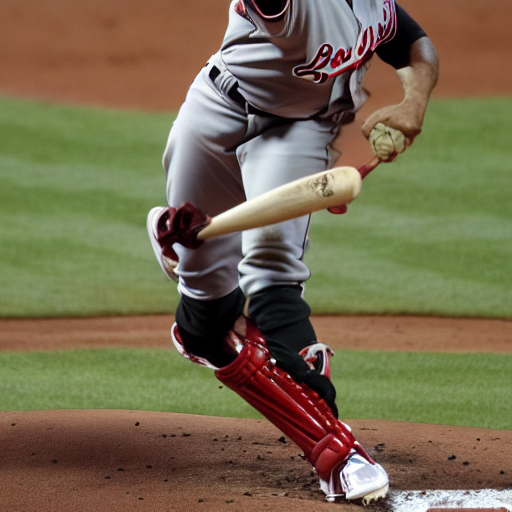}
&\includegraphics[width=0.15\linewidth]{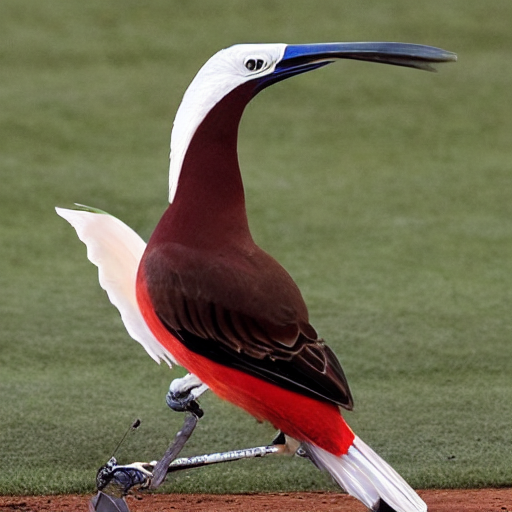}
&\includegraphics[width=0.15\linewidth]{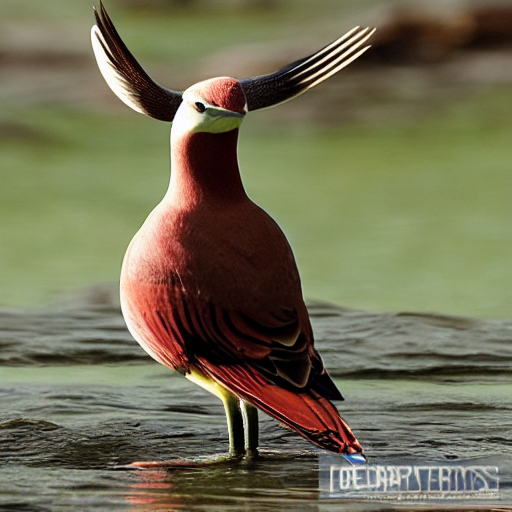}
&\includegraphics[width=0.15\linewidth]{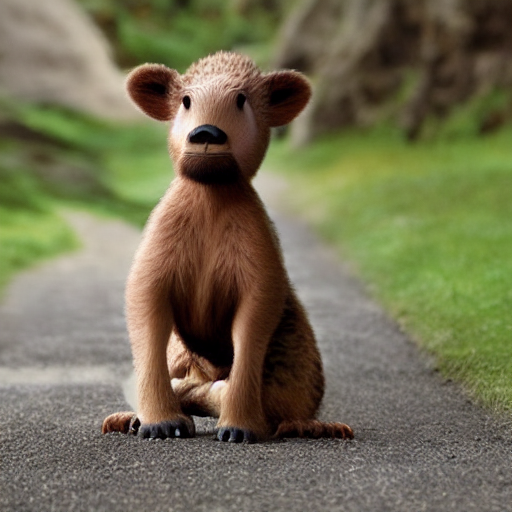}
&\includegraphics[width=0.15\linewidth]{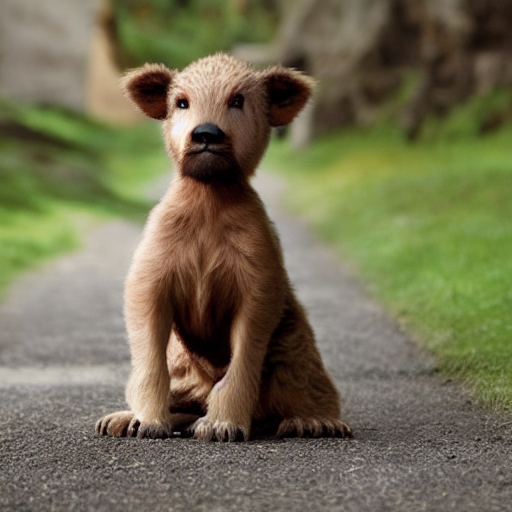}
&\includegraphics[width=0.15\linewidth]{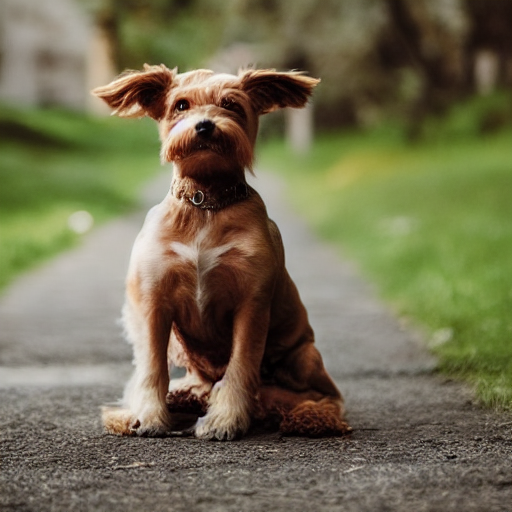}
&\includegraphics[width=0.15\linewidth]{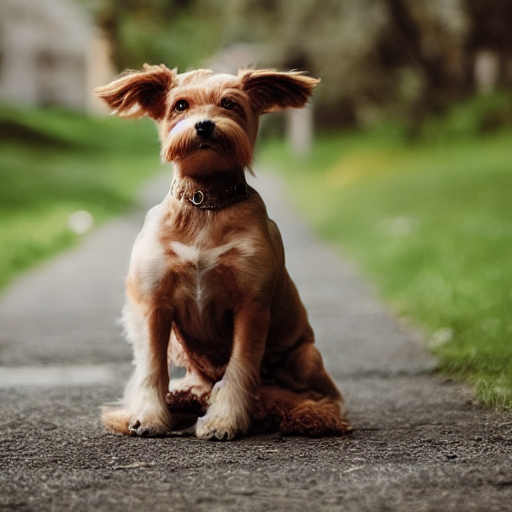}\\
\includegraphics[width=0.15\linewidth]{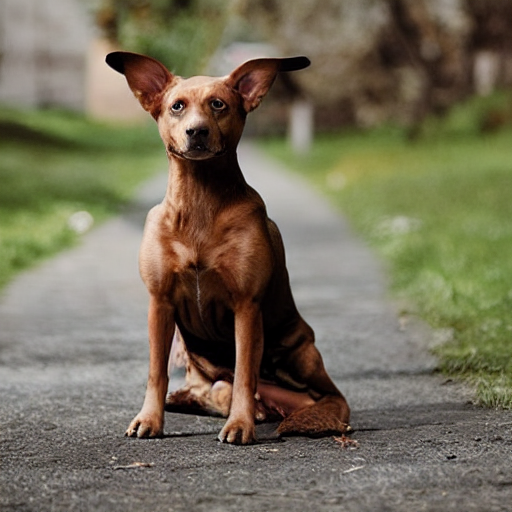}
&\includegraphics[width=0.15\linewidth]{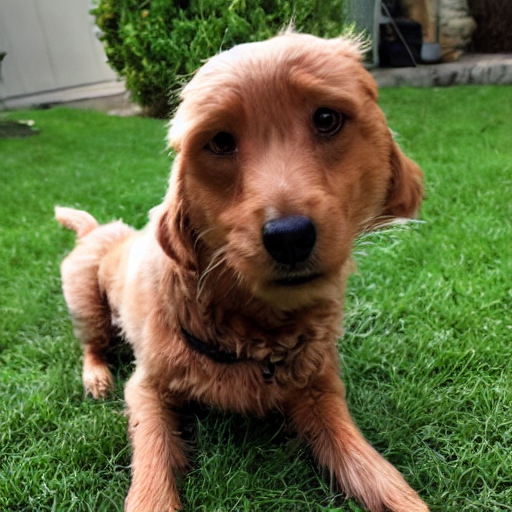}
&\includegraphics[width=0.15\linewidth]{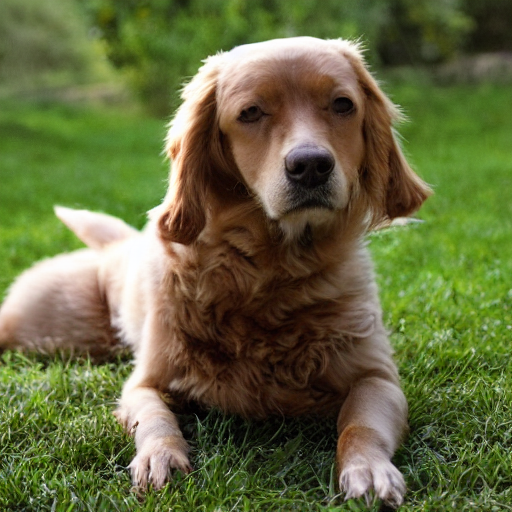}
&\includegraphics[width=0.15\linewidth]{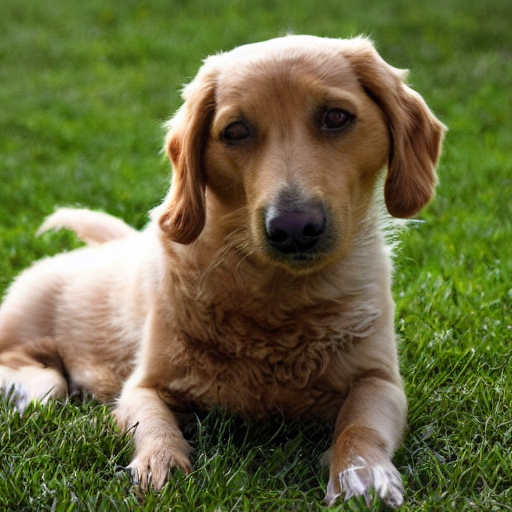}
&\includegraphics[width=0.15\linewidth]{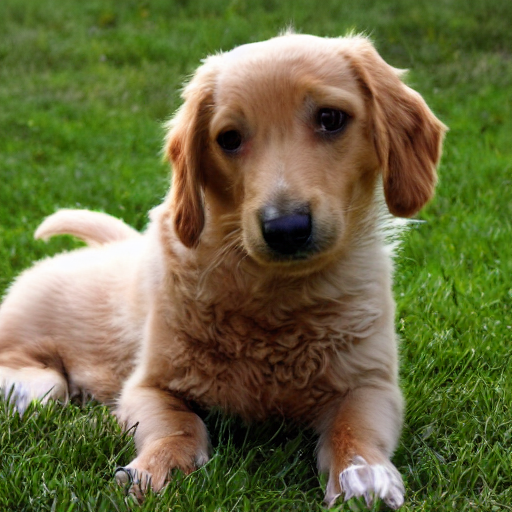}
&\includegraphics[width=0.15\linewidth]{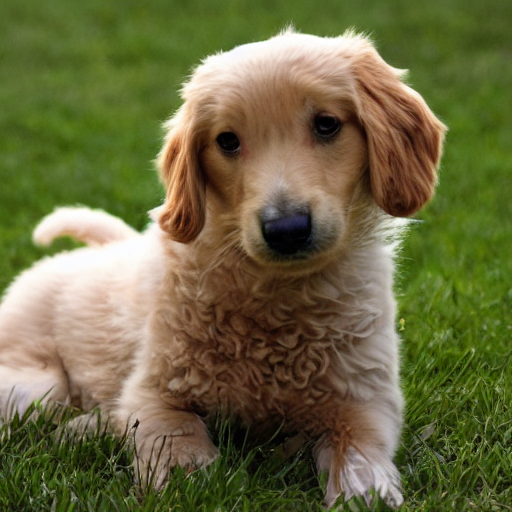}
&\includegraphics[width=0.15\linewidth]{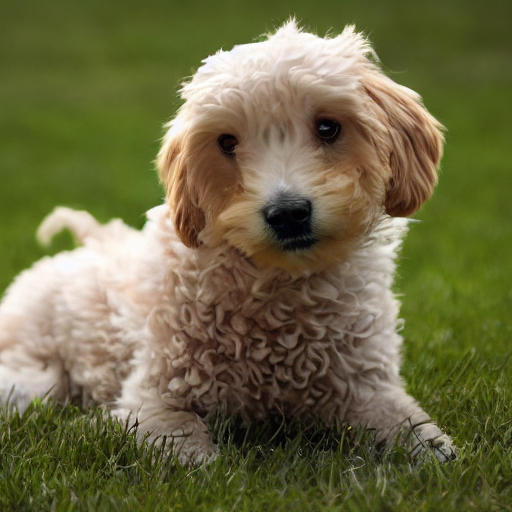}
&\includegraphics[width=0.15\linewidth]{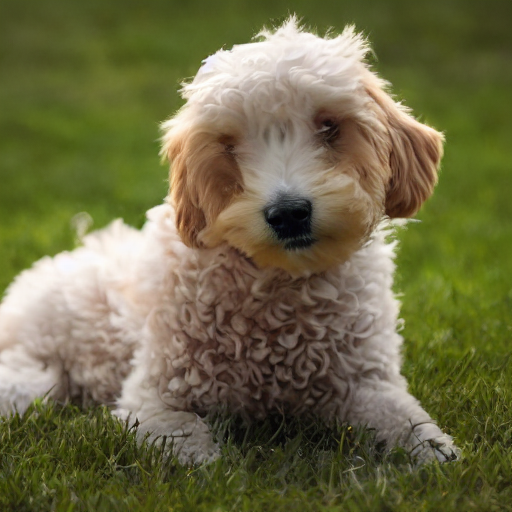}
&\includegraphics[width=0.15\linewidth]{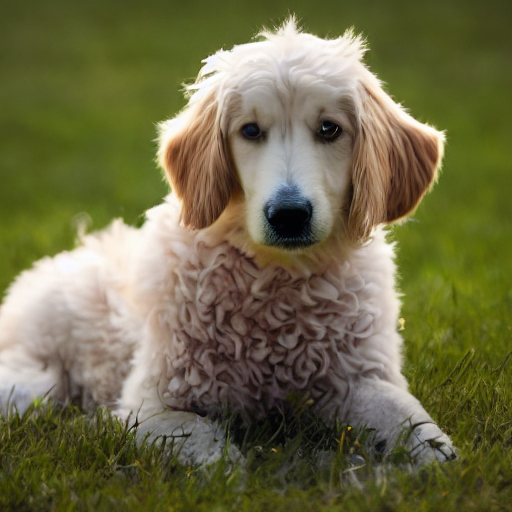}
&\includegraphics[width=0.15\linewidth]{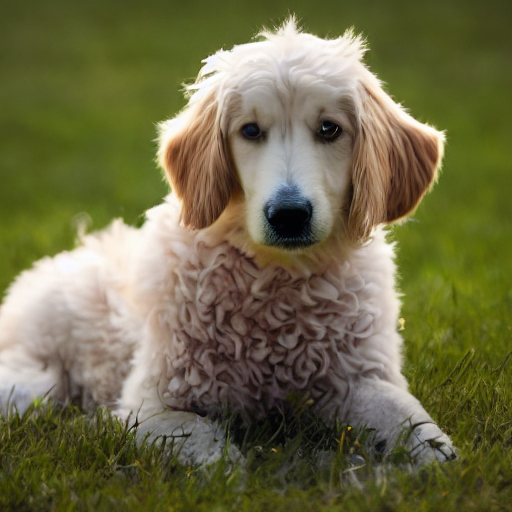}
\\[-0.07cm]
\end{tabular}}
\vspace{-0.12in}
\caption{\textbf{Top:} Negative images produced by SD (T = 40 step) using \dns for the latent $z_t$ of the \cfg chain, for the prompt {\bf p} = {\tt photo of a dog}, at different time steps $t$. \textbf{Bottom:} \ours images corresponding to $t$ \ours steps and remaining \cfg steps.}
\label{fig:chng_neg_imgs}
\end{figure*}

\noindent\textbf{Classifier-free Guidance (\cfg)~\cite{cfg}:} provides a trade-off between diversity and compliance by training the model both with (90\%) and without (10\%) the prompt $\bf p$. A linear combination of prompt-conditional $(\hat{\epsilon}_p)$ and unconditional ($\hat{\epsilon}_\phi$) noise estimates is used at inference time.
\begin{equation}\label{eq:diff_inf}
    \hat{\epsilon} = \hat{\epsilon}_\phi + s(\hat{\epsilon}_p - \hat{\epsilon}_\phi),
\end{equation}
where $\hat{\epsilon}_p= \epsilon_\theta(z_t;t,\tau({\bf p}))$, $\hat{\epsilon}_\phi= \epsilon_\theta(z_t;t,\tau(\phi))$, and $s$ is known as the guidance-scale that controls the conditioning strength. At each step, the noise $\hat{\epsilon}$ is used to denoise the latent. Under the energy-based interpretation, this is equivalent to sampling from
\begin{equation}
    \begin{aligned}\label{eq:cfg_p}
         p_{\textit{cfg}}(z_t, {\bf p}) \propto p_\theta(z_t) p_\theta({\bf p} |z_t)^s.
    \end{aligned}
\end{equation}
While increasing $s$ can strengthen the conditioning process, it cannot offset an extremely low  $p_\theta({\bf p} |z_t)$. This is shown in \Cref{fig:samples}, where \cfg fails even for high guidance values.

\noindent\textbf{Negative Prompting (\np)~\cite{compos}:} is a popular tool in the DM user community to improve prompt compliance and image quality. It leverages a negative prompt $\bf n$ to provide additional guidance to the DM. Most systems~\cite{AUTOMATIC1111} implement negative prompting in a slightly different manner than introduced by \cite{compos}, replacing~\cref{eq:diff_inf} by the denoising equation
\begin{equation}\label{eq:negp_emb}
    \hat{\epsilon} = \hat{\epsilon}_{n} + s(\hat{\epsilon}_p -\hat{\epsilon}_{n} )
\end{equation}
where $\hat{\epsilon}_{n} = \epsilon_\theta(z_t;t,\tau({\bf n}))$. Under the energy-based interpretation, this corresponds to sampling from 
\begin{equation}
    \begin{aligned}\label{eq:np_p}
         p_{\textit{np}}(z_t, {\bf p}) \propto p_\theta(z_t) {\left[\frac{p_\theta({\bf p} |z_t)}{p_\theta({\bf n} |z_t)}\right]}^s.
    \end{aligned}
\end{equation}

\noindent\textbf{Diffusion-Negative Prompting (\dnp)~\cite{dnp}:} seeks the optimal negative prompt. It notes that \cref{eq:np_p} encourages the sampling of latent codes $z_t$ of large odds ratio
\begin{equation}
    o(z_t, {\bf p}, {\bf n}) = \frac{p_\theta({\bf p}| z_t)}{p_\theta( {\bf n}| z_t)}.
    \label{eq:oddr}
\end{equation}
 Since the  Bayes decision rule for deciding between ${\bf p}$ and ${\bf n}$ is to choose ${\bf p}$ when $o(z_t, {\bf p}, {\bf n}) \ge 1$ and ${\bf n}$ otherwise, this increases the probability that sampled codes comply with prompt ${\bf p}$ and not with prompt ${\bf n}$. It then defines the optimal negative prompt ${\bf n^*}$ {\it for the DM\/} as the one that maximizes the steepness of the odds ratio with $\bf p$, i.e.
\begin{equation}
    {\bf n}^*({\bf p}) = \arg\max_{\bf n} \nabla_{z_t} \log o(z_t, {\bf p}, {\bf n}). \label{eq:opt}
\end{equation}
The authors of \cite{dnp} show that this leads to the {\it Diffusion Negative Sampling\/} (\dns) equation
\begin{equation}\label{eq:diff_inf_neg}
    \hat{\epsilon} = \hat{\epsilon}_\phi +s(\hat{\epsilon}_\phi - \hat{\epsilon}_p)
\end{equation}
for sampling optimal negative images. \dnp samples one such image and captions it to estimate the optimal negative prompt, $\bf n^*$. This is finally used in the \np equation of \cref{eq:negp_emb} to synthesize the desired image.

Both guidance scale and negative prompting can increase the odds ratio in \cref{eq:np_p}. However, the difference between them can be seen by comparing \cref{eq:cfg_p} and \cref{eq:np_p}. In \cref{eq:cfg_p}, increasing the guidance scale does not work well when $p_\theta({\bf p}| z_t)$ is very low due to the seed choice. Given the same seed, we can use \cref{eq:opt} to estimate the ``optimal" negative prompt that can maximize the odds ratio and increase $p_{\textit{np}}(z_t, {\bf p})$. Another difference is observed in practice, where increasing the guidance scale beyond a certain threshold causes the noise to be outside the distribution of the trained DM and can, therefore, result in over-saturation or artifacts. This can be avoided with negative prompting because the noise distribution remains unchanged.

\section{Proposed Method}

\subsection{Sampling Procedures}
\Cref{fig:overview} shows the different sampling procedures discussed in this work. \cfg uses a single Markov chain conditioned by prompts $\bf p$ and $\phi$ to produce the denoised latent $z_0$. \dnp uses two, a \dns chain to estimate ${\bf n^*}$ and another to generate the final image using $({\bf p}, {\bf n^*})$. 
\begin{equation}\label{eq:DNP_n_est}
{\bf n^*}({\bf p}) = {\cal C}(\text{DNS}_{k=T}(z_T, {\bf p}))
\end{equation}
where $\text{DNS}_{k}$ denotes $k$ \dns steps and $\cal C$ is an external captioning model. The prompt ${\bf n^*}({\bf p})$ is then used, instead of $\phi$, in the \np chain to generate the final image. This has a few limitations. First, $\cal C$ can be much larger than the DM, increasing complexity. Second, much of the information in the negative image synthesized by \dns is ``lost in translation". A caption is, by definition, a limited representation of image content, and $\cal C$ may not even pay attention to the visual details that make the image a negative from the DM perspective. Finally, and most importantly, the fundamental \dnp assumption that a single caption  ${\bf n^*}({\bf p})$ is optimal for every diffusion step $t$, in~\cref{eq:negp_emb}, is questionable. This can be observed simply by inspection of~\cref{eq:oddr,eq:opt}. Since the latent $z_t$ varies with $t$, so will the optimal prompt ${\bf n^*}({\bf p})$.

\subsection{Complete diffusion-Negative Prompting~(\cnp)}

In this work, we hypothesize that this assumption does not hold, i.e., \dns produces a different negative image for the latent $z_t$ of each step $t$ and, consequently,
\begin{equation}
\text{DNS}_{k=T}(z_T, {\bf p}) \neq \text{DNS}_{k=t} (z_{t}, {\bf p})~~\forall~ t \in [0, T). \label{eq:DNSnew}
\end{equation}
To test this hypothesis, we used the \dns chain of~\cref{eq:diff_inf_neg} to generate the diffusion-negative image $I_{n,t}$ for different latents $z_t$, sampled with \cfg for a given prompt ${\bf p}$ and latent $z_0$. The top half of \Cref{fig:chng_neg_imgs} shows the image $I_{n,t}$ for various values of $t$.  As the Markov chain of \cref{eq:diff_inf} progresses and $z_t$ changes, the corresponding diffusion-negative image $I_{n,t}$ and corresponding optimal negative prompt, ${\bf n^*}({\bf p})$ change as well.  For example, the negative image for ``a photo of a dog" depicts a living room, cyclist, baseball player, birds, and dogs as $t$ varies. To reflect this dependence clearly, we modify the notation from ${\bf n^*}({\bf p})$ to ${\bf n^*}({\bf p},t)$. This is also more consistent with \cref{eq:oddr,eq:opt} and leads to
\begin{equation}\label{eq:CNS}
{\bf n^*}({\bf p},t) = {\cal C}(\text{DNS}_{k=t}(z_t, {\bf p}))
\end{equation}
Comparing with \cref{eq:DNP_n_est} shows that the prompt ${\bf n^*}$ of \dnp can be written as ${\bf n^*}({\bf p},T)$ in the new notation and is optimal only for the first \dnp step ($t=T$). 

To ensure optimality at every step, ${\bf n^*}({\bf p},t)$ must be estimated for every $t \in (0, T]$. This is denoted {\it Complete diffusion-Negative Prompting\/} (\cnp). As illustrated in \Cref{fig:overview}, it consists of implementing~\cref{eq:CNS} instead of \cref{eq:DNP_n_est} by initializing the \dns chain at $z_t$ and running it for $t$ steps, at each step of the \np chain. The resulting negative prompt ${\bf n^*}({\bf p},t)$ is then used in the next step of the \np chain, ensuring that the negative prompt used at every step is optimal for that step. 

\noindent\textbf{Limitations of \cnp:} While \cnp eliminates \dnp's limitation of using a single negative for all steps, the added complexity is substantially larger because \cref{eq:CNS} requires using $\cal C$ at every time step $t$. As $t$ approaches $0$ (end of the chain), only small details are edited and $\cal C$ may ignore the details that differentiate the positive and negative conditions. For example, as shown in the top half of  \Cref{fig:chng_neg_imgs} for ${\bf p}=\text{``a photo of a dog"}$, as $t$ approaches $0$, the diffusion negative image also contains a dog (low quality, distorted, etc.). The correct negative prompt should be ${\bf n^*}({\bf p},t) = \text{``a distorted dog"}$, to improve the image quality. However, captioning models caption it as ``dog", which is detrimental. Beyond this, the complexity of \cnp is $O(T^2)$, which is intractable for most applications. Hence, while better justified mathematically, in practice, \cnp can be intractable and more prone to captioning errors. 

\subsection{\textbf{\ours}}

\begin{algorithm}[t]
\SetAlgoLined
    \caption{\ours}
    \textbf{Input}: DM $\epsilon_\theta$, text prompt ${\bf p}$, initial noise, $z \sim\mathcal{N}(0,I)$, number of timesteps $T$, number of \dns timesteps $K$, guidance scale $s$, negative guidance scale $s_{n}$, scheduler $Schdlr$\\
    
    \For{$t \gets T$ \KwTo $1$}{
        $\epsilon^{t}_p \gets \epsilon_\theta(z, {\bf p}, t), \epsilon^{t}_\phi \gets  \epsilon_\theta(z,``~", t)$ \;

        $K_t \gets Schdlr(K, t)$ \;
        \If{$K_t > 0$ \textit{and} $t > T/2$}{
            $z^{n} \gets z$ \;
            \For{$i_{n} \gets 0$ \KwTo $K_t-1$}{
                $t_{n} \gets t$ - $i_{n}$ \;
                
                $\epsilon^{t_{n}}_p \gets \epsilon_\theta(z^{n}, {\bf p}, {t_{n}})$ \;
                
                $\epsilon^{t_{n}}_\phi \gets \epsilon_\theta(z^{n}, ``~", {t_{n}})$ \;
                 $\epsilon^{t_{n}} \gets \epsilon^{t_{n}}_p +s_{n}(\epsilon^{t_{n}}_\phi-\epsilon^{t_{n}}_p)$ \tcp{DNS;}
                 
                 $z^{n} \gets \textit{Sample}(z^{n}, \epsilon^{t_{n}}, {t_{n}})$ \;
            }
            $\epsilon^{t}_{n} \gets \textit{Normalize}(\epsilon^{t_{n}}, t)$ \;
            
        
            $\epsilon^{t} \gets \epsilon^{t}_{n} +s(\epsilon^{t}_p -\epsilon^{t}_{n})$ \tcp{NP;}   
        }
        \Else{
            $\epsilon^{t} \gets \epsilon^{t}_\phi +s(\epsilon^{t}_p -\epsilon^{t}_\phi)$ \tcp{CFG;}
        }
    
        $z \gets \textit{Sample}(z, \epsilon^{t}, {t})$ \;
    }\label{algo:ANS}
\end{algorithm}

\begin{figure*}[t]\RawFloats
\noindent \centering
\begin{minipage}[t]{0.6\textwidth} 
    \textbf{}\\[1pt]
    \resizebox{\linewidth}{!}{
        \begin{tabular}{ccccccc}
        \hline
        \multirow{2}{*}{\textbf{Dataset}} & \multirow{2}{*}{\textbf{Method}} & \multirow{2}{*}{\textbf{CLIP}} & \multirow{2}{*}{\textbf{IS}} & \textbf{Image}  & \textbf{Pick} & \multirow{2}{*}{\textbf{HPSv2}}\\
         & & & & \textbf{Reward} &  \textbf{Score} &  \\
        \hline\hline
        
        \multirow{3}{*}{\textbf{Attend\&Excite}} 
        & SDXL~(\cfg)        & 32.97 (27\%) & 11.24 & 0.95 (26\%) & 22.73 (33\%) & 30.07 (24\%) \\
        & SDXL+\dnp          & 33.20 (21\%) & \textbf{11.65} & 1.03 (25\%) & 22.79 (26\%) & 30.38 (26\%) \\
        & SDXL+\ours         & \textbf{33.62} \textbf{(52\%)} & 11.15 & \textbf{1.20} \textbf{(49\%)} & \textbf{22.88} \textbf{(41\%)} & \textbf{30.92} \textbf{(50\%)} \\
        \hline
        
        \multirow{3}{*}{\textbf{Pick-a-Pic}} 
        & SDXL~(\cfg)        & 33.42 (27\%) & 11.11 & 0.91 (24\%) & 22.14 (29\%) & 30.24 (24\%) \\
        & SDXL+\dnp          & 33.47 (21\%) & \textbf{11.67} & 0.85 (25\%) & 22.08 (23\%) & 30.47 (25\%) \\
        & SDXL+\ours         & \textbf{34.09} \textbf{(52\%)} & 11.28 & \textbf{1.00} \textbf{(51\%)} & \textbf{22.39} \textbf{(48\%)} & \textbf{30.95} \textbf{(51\%)} \\
        \hline
        
        \multirow{3}{*}{\textbf{DrawBench}} 
        & SDXL~(\cfg)        & 32.59 (28\%) & 14.88 & 0.61 (28\%) & 22.37 (32\%) & 28.86 (24\%) \\
        & SDXL+\dnp          & 32.87 (22\%) & \textbf{15.34} & 0.71 (26\%) & 22.45 (24\%) & 29.38 (30\%) \\
        & SDXL+\ours         & \textbf{33.27} \textbf{(50\%)} & 15.19 & \textbf{0.79} \textbf{(46\%)} & \textbf{22.56} \textbf{(44\%)} & \textbf{29.45} \textbf{(48\%)} \\
        \hline
        
        \multirow{3}{*}{\textbf{PartiPrompts}} 
        & SDXL~(\cfg)        & 32.33 (28\%) & 16.35 & 0.65 (26\%) & 22.27 (33\%) & 28.43 (23\%) \\
        & SDXL+\dnp          & 32.02 (22\%) & 16.45 & 0.62 (26\%) & 22.18 (23\%) & 28.72 (31\%) \\
        & SDXL+\ours         & \textbf{32.91} \textbf{(50\%)} & \textbf{16.55} & \textbf{0.81} \textbf{(48\%)} & \textbf{22.39} \textbf{(44\%)} & \textbf{28.97} \textbf{(47\%)} \\
        \hline
        
        \end{tabular}
        }
        \captionof{table}{Quantitative results on the Attend\&Excite, Pick-a-Pic, DrawBench, and PartiPrompts datasets. \%WinRates are shown in parentheses next to each metric. Best values are in \textbf{bold}.}\label{tab:pkdb}
    \resizebox{\linewidth}{!}{
        \begin{tabular}{cccccccc}
        \hline
        \multirow{2}{*}{\textbf{Dataset}} & \multirow{2}{*}{\textbf{Method}} & \multirow{2}{*}{\textbf{CLIP}} & \multirow{2}{*}{\textbf{IS}} & \multirow{2}{*}{\textbf{FID}}  & \textbf{Image}  & \textbf{Pick} &\multirow{2}{*}{\textbf{HPSv2 }}\\
         & & &&&  \textbf{Reward} & \textbf{Score} & \\
          \hline\hline
         \multirow{3}{*}{\textbf{ImageNet}}&SDXL~(\cfg) & 30.22 & 27.26 & 69.8  & 0.21 & 21.24 & 25.63\\
         &SDXL+\dnp & 30.19 & 34.55 & 70.11  & 0.26 & 21.30& 26.80\\
        &SDXL+\ours & \textbf{30.64} & \textbf{40.83} & \textbf{67.67}  & \textbf{0.48} & \textbf{21.46} & \textbf{26.87}\\
        
        \hline
        \end{tabular}}
        \captionof{table}{Quantitative Results on the Imagenet dataset. Best results are in \textbf{Bold}.}
        \label{tab:im}
\end{minipage}%
\hfill
\begin{minipage}[t]{0.39\textwidth} 
        
    \vspace{-0.28in}
    \textbf{}\\[1pt]

    \includegraphics[width=\linewidth]{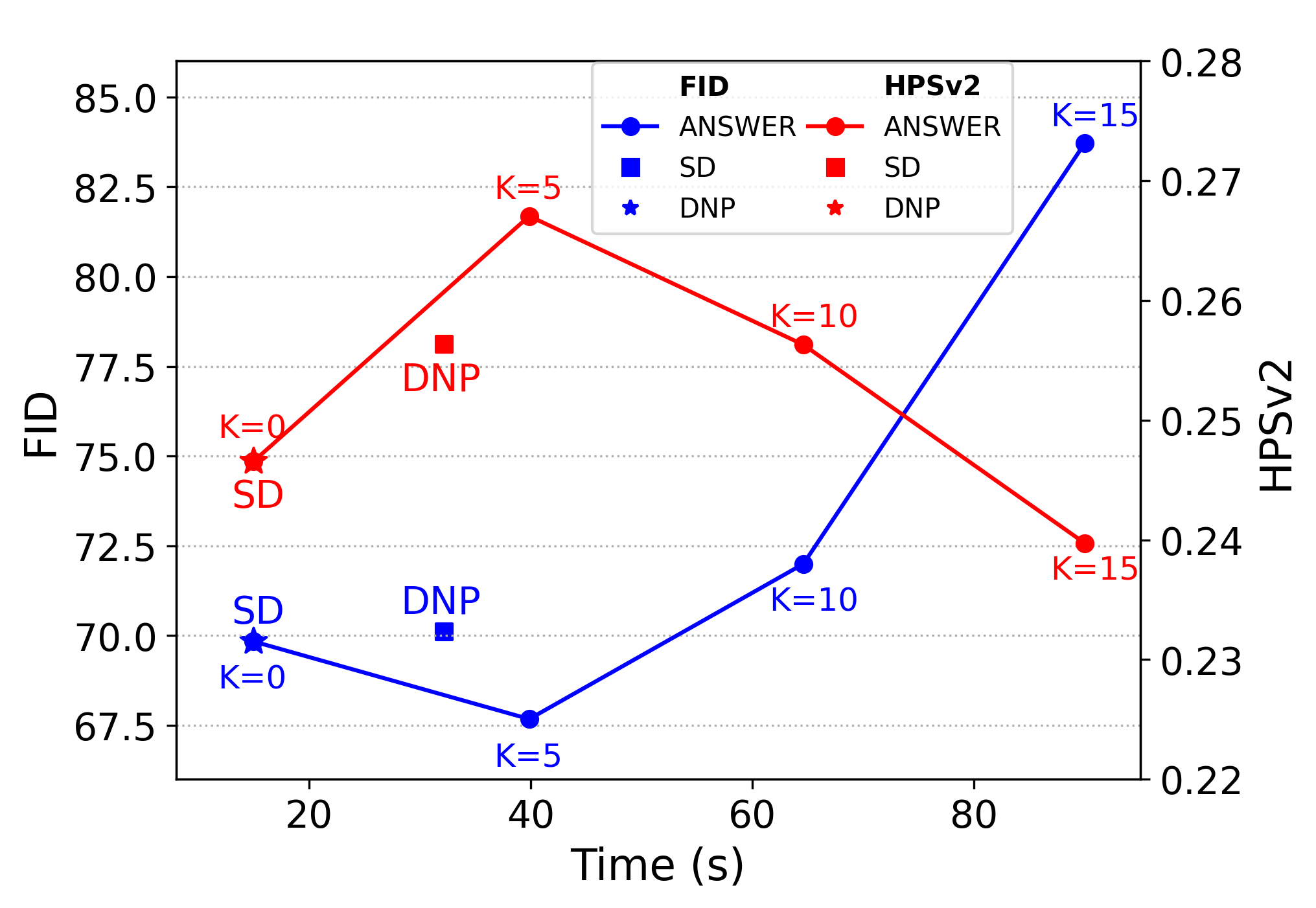}
     \caption{Inference time Pareto curves over K: FID (Left axis) and HPSv2 (Right axis)}
    \label{fig:abls_k}

     \resizebox{\linewidth}{!}{
        \begin{tabular}{c@{\hskip 0.08em}c@{\hskip 0.12em}c@{\hskip 0.08em}c}
        \multicolumn{2}{c}{{\bf p} = {\tt Digital illustration, Burger }} & \multicolumn{2}{c}{{\bf p} = {\tt Dog and Santa. Black and white }} \\
        \multicolumn{2}{c}{{\tt with wheels on the race track}} & \multicolumn{2}{c}{{\tt Christmas trees in background}}\\
        \includegraphics[width=.54\linewidth]{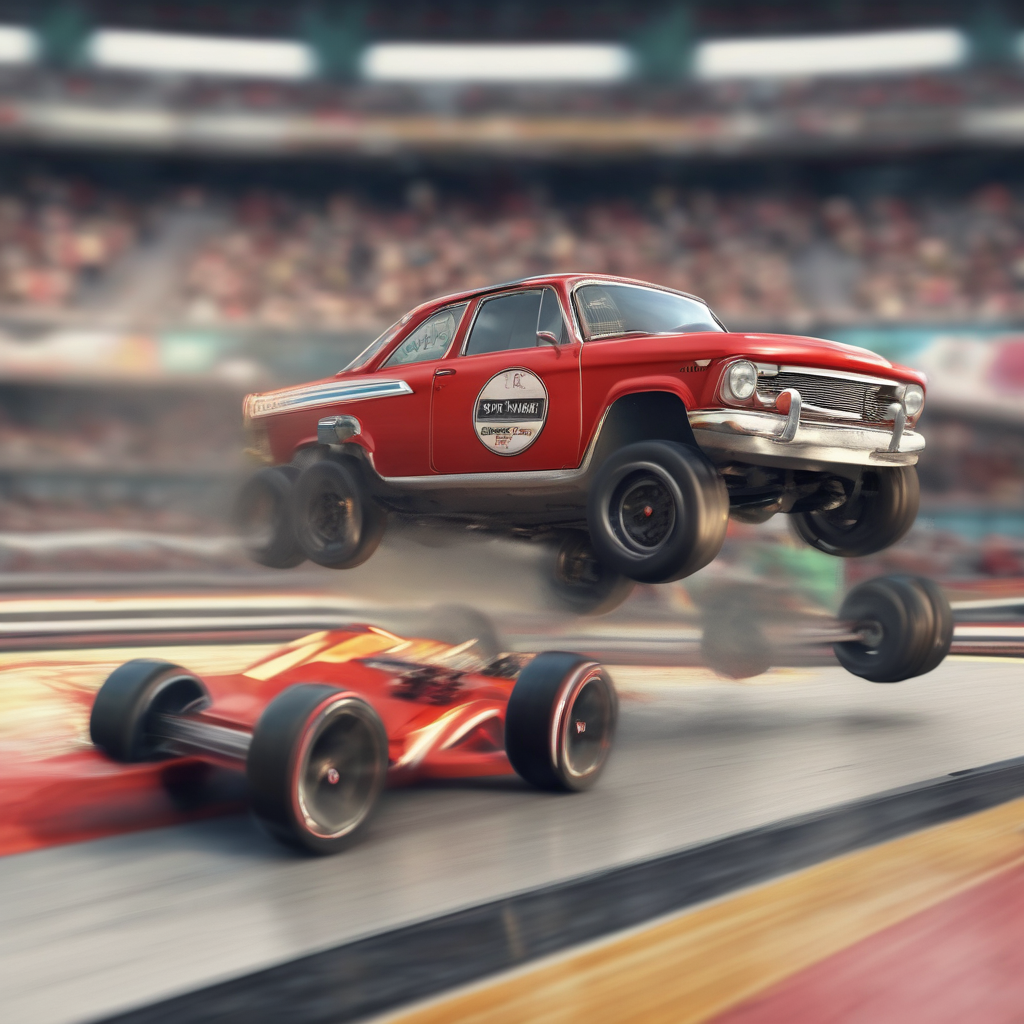} &\includegraphics[width=.54\linewidth]{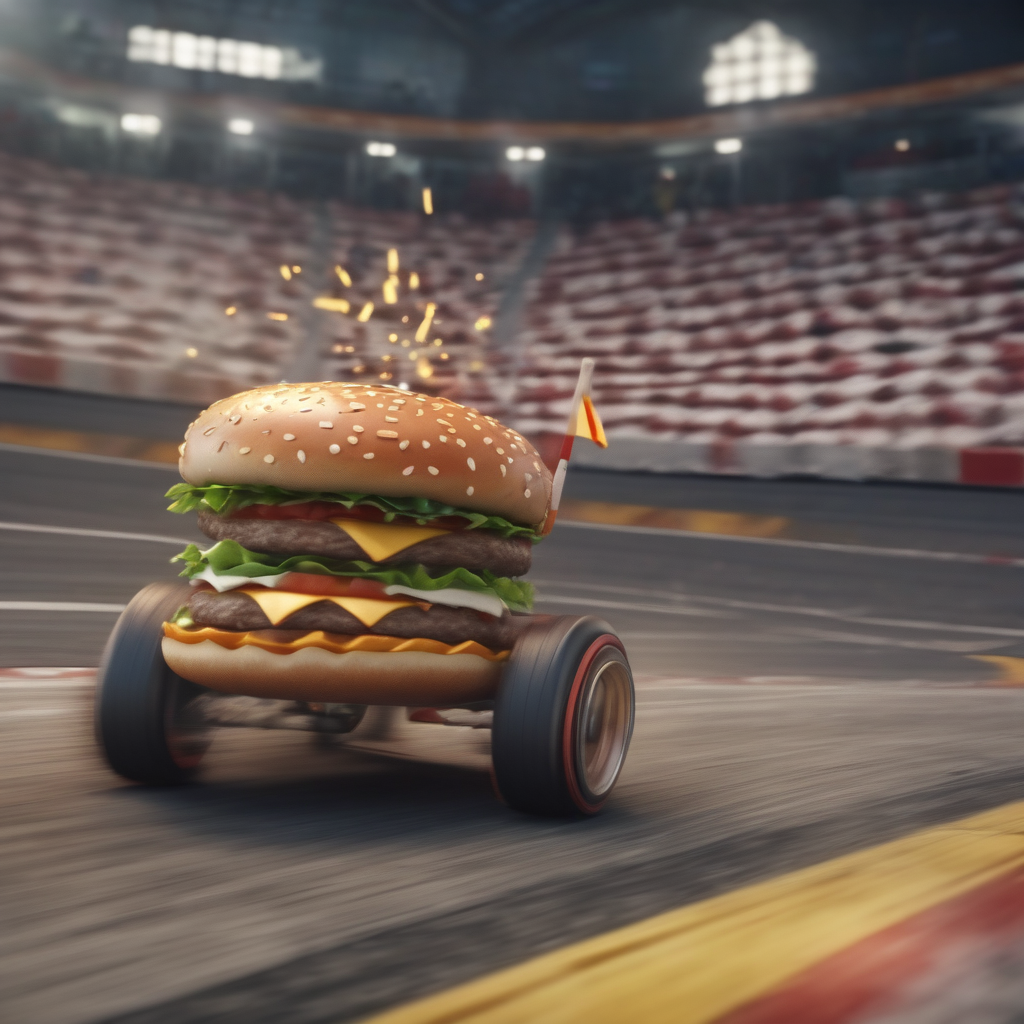} &

        \includegraphics[width=.54\linewidth]{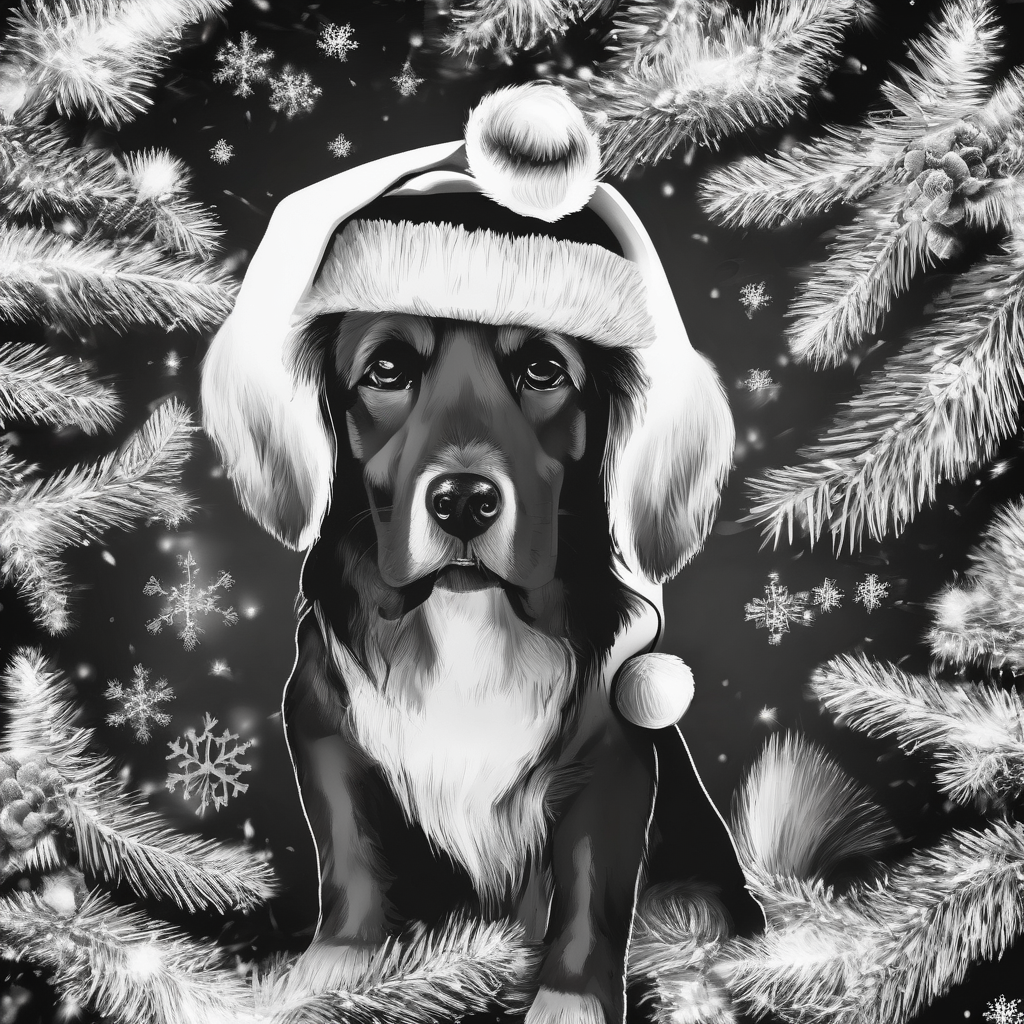} &
        \includegraphics[width=.54\linewidth]{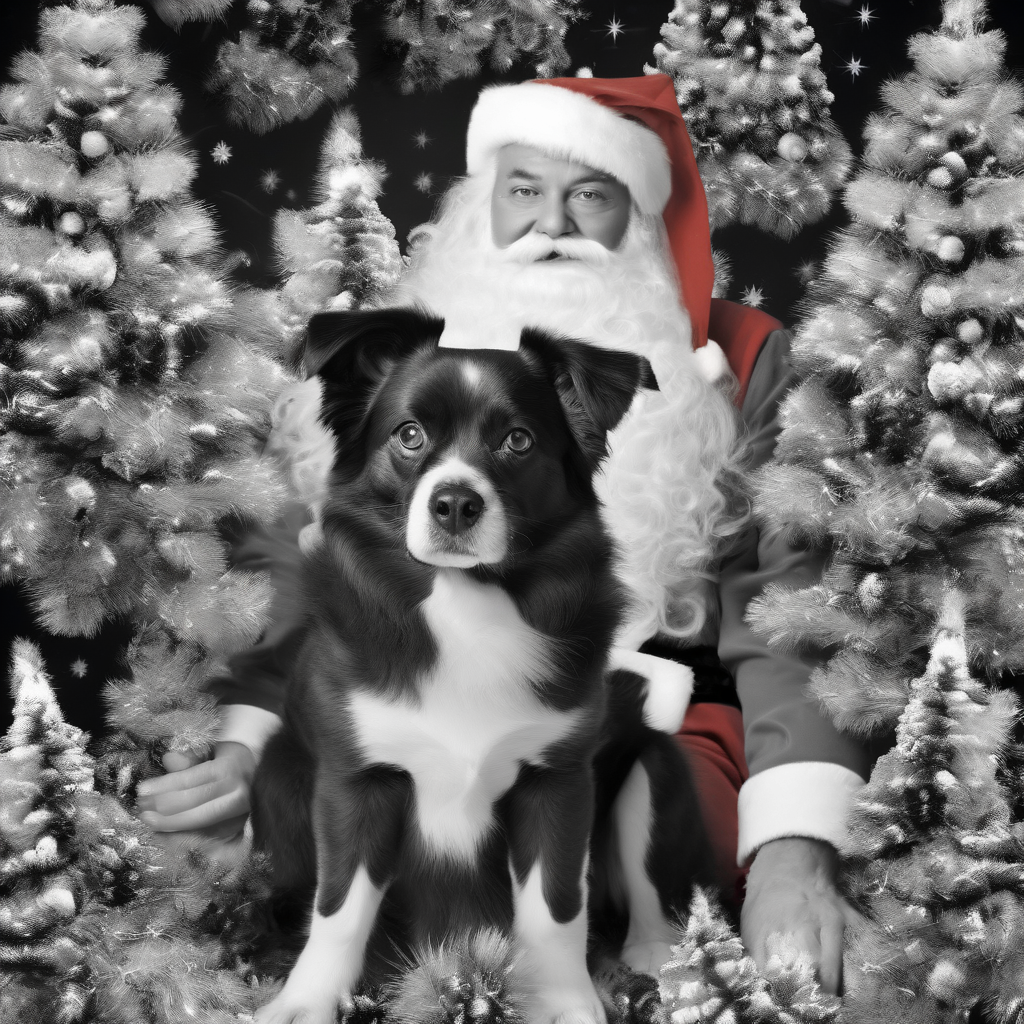}  \\
        \end{tabular}}
    \caption{Complex and imaginary scenarios: (Left: SDXL~(\cfg), Right: SDXL+\ours)}
    \label{fig:more_result}

    \end{minipage}
\end{figure*}

To address these problems, we propose the  {\it Adaptive Negative Sampling Without External Resources\/} (\ours) sampling procedure of \Cref{fig:overview}. This is a variant of \cnp that eliminates the external captioning model. At step $t$, rather than producing a negative caption ${\bf n^*}({\bf p},t)$ and using it to condition the noise term $\hat{\epsilon}_{n}$ of~\cref{eq:negp_emb}, this noise term is replaced by an estimate produced by running a few ($K_t$) steps of the \dns chain, namely the negative noise produced by this chain at time $t-K_t$. The intuition is that the solution of \cref{eq:opt} is actually the term $-\hat{\epsilon}_{p}$ of~\cref{eq:diff_inf_neg}. The remaining  $\hat{\epsilon}_{\phi}$ terms result from a normalization to produce a valid probability distribution and some arguments regarding the need to always define negative conditions in context~\cite{compos_prev} (see~\cite{dnp} for details). While for \dns that context is the distribution of natural images, associated with prompt $\phi$, the \np chain of \cref{eq:negp_emb} already establishes a context: the noise distribution $\hat{\epsilon}_{p}$ associated with the positive prompt $\bf p$. Note that if the \dns chain were run for a single iteration ($K_t=1$) and the same denoising network used in the \dns and \np chains, \cref{eq:negp_emb} would reduce to $\hat{\epsilon} = \hat{\epsilon}_\phi + (s^2-s+1)(\hat{\epsilon}_p - \hat{\epsilon}_\phi)$, which is equal to simply increasing the guidance scale of \cfg. While this effectively increases sampling odds, $\epsilon_{n^*}$ remains $\epsilon_\phi$, and the Markov chain is not substantially changed. Using a small $K_t>1$ allows $\epsilon_{n^*}$ to diverge from $\epsilon_\phi$ and further increase the odds ratio of~\cref{eq:np_p}. This also makes the sampling highly adaptive to the prompt $\bf p$ and can be implemented with no external resources, neither in the form of a captioning model nor a user-specified negative prompt.

An implementation of \ours is detailed in Algorithm~\ref{algo:ANS}. At diffusion step $t$, $K_t$ iterations of \dns sampling are implemented with \cref{eq:diff_inf_neg}. $K_t$ is the number of \dns timesteps and is determined by a scheduler, $Schdlr(K,t) = K \cdot\left( \frac{2t - T}{2t - T + 20} \right) \cdot \left( \frac{T + 20}{T} \right)$, where $T$ is the total number of timesteps and $t \in \{T,..,T/2\}$ is the current timestep. This gradual decay, starting at $K_T=K$, allows a smoother transition between negative and unconditional noise. The negative noise $\epsilon^{t_n}$ is normalized to match the noise statistics $\epsilon^t_\phi$ using  $ Normalize(\epsilon^{t_n},t) = \left( \frac{\epsilon^{t_n} - \mu(\epsilon^{t_n})}{\sigma(\epsilon^{t_n})}\right) \cdot \sigma(\epsilon^t_\phi) + \mu(\epsilon^t_\phi)$. This is helpful because the noise distributions change with time step $t$, and the negative noise is $K_t$-steps ahead. The normalized noise $\epsilon^t_n$ is then passed on to the \np step of \cref{eq:negp_emb}. Experimentally, this negative prompting step is more effective for early time steps $t$, where the image is still poorly formed but does not help in the later steps. For example, the bottom half of \Cref{fig:chng_neg_imgs} shows the images created by performing $t$ \ours steps followed by the $T-t$ \cfg steps. Therefore, applying \ours beyond a certain number of steps in the \np chain does not increase performance significantly. In this example, the negative and positive chain converge around $t=12$, and the positive image does not change much beyond that point. Additionally, latent editing literature~\cite{ae,syngen} observes that only cosmetic changes can be made in the final timesteps. Therefore, to save time and resources, we replace the \np chain of \cref{eq:negp_emb} with the standard \cfg chain for $t < T/2$. Restricting to the early stages also reduces inference time of \ours. Stopping criteria can be researched, as different prompt/seed pairs may need more or fewer steps based on difficulty.


Note that, rather than operating in the text domain, \ours defines the negative prompt directly in noise space. This eliminates the ``lost in translation" problem associated with captions. Since \ours does not estimate the negative prompt at each step, it does not need to complete the Markov chain and can use any $\epsilon^{t_n}$ from the DNS chain as a substitute for the negation. This allows choosing the number of steps $K$, thereby reducing the computational complexity of \ours to $O(KT)$, which is tractable for low values of $K$.  Overall, \ours captures dynamic changes in ${\bf n^*}({\bf p},t)$ through each step without requiring full sampling to the final step. As a result, sampling with \ours is tractable, independent of external models, and estimates an adaptive negative noise for every step in the diffusion chain. Using the negative noise obtained from \dns in the diffusion process bridges the semantic gap more effectively. This also allows \ours to capitalize on the flexibility of \dns sampling to improve efficiency and effectiveness for longer or more complex prompts.

\vspace{-0.15in}
\begin{figure*}[t]
\centering 
\includegraphics[width=\linewidth]{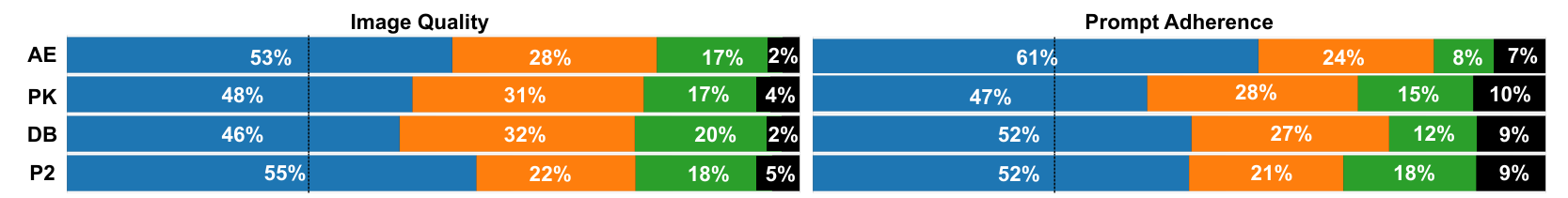}
\vspace{-0.25in}
\caption{\textbf{Human evaluation using Amazon Mechanical Turk:} From left to right: SDXL+\ours is denoted by {\color{barblue} \textbf{blue}},  SDXL+\dnp denoted by {\color{barorange} \textbf{orange}}, SDXL~(\cfg) is denoted by {\color{bargreen} \textbf{green}} and {\it `No Clear Winner'} is {\color{black} \textbf{black}}.}
\label{fig:humaeval}
\end{figure*}

\section{Experiments}
We discuss the experiments we performed to evaluate \ours in this section. Comprehensive details about datasets and evaluation metrics with additional results can be found in the Supplementary.

\noindent \textbf{Datasets:} We evaluate \ours using five prominent datasets to assess its ability to generate images with comprehensive and diverse descriptions that extend beyond common training data. Specifically, we utilize the ImageNet (IM)~\cite{ILSVRC15}, Attend \& Excite (AE)~\cite{ae}, Pick-a-Pic (PK)~\cite{pick}, DrawBench (DB)~\cite{imagen}, and PartiPrompts (P2)~\cite{parti} datasets. These datasets test the model's ability to handle entity attributes and interactions, spatial relationships, text rendering, numeracy, and rare or imaginative scenarios. 

\noindent\textbf{Evaluation metrics:} We use both human and automated evaluation metrics for the large-scale assessment of the generated images. The CLIP Score~\cite{clipscore}, FID~\cite{fid} and Inception Score (IS)~\cite{IS} are used as the traditional measures along with emerging popular Human Preference Metrics, such as HPSv2~\cite{hpsv2}, ImageReward~\cite{reward} and PickScore~\cite{pick}. Since, they are trained as preference metrics, not for absolute scoring, their absolute values are not highly informative of visual quality. Since, small metric deltas can correspond to large differences in visual quality, we also report win-rate percentages to offer further insights into generation quality.

\subsection{Quantitative Results}

We evaluate \ours on Stable Diffusion (SD) and Stable Diffusion XL (SDXL). \Cref{tab:pkdb} compares the performance of models using \ours to the respective baselines and their negative prompting with \dnp, on the AE dataset, Pick-a-Pic, DrawBench, and PartiPrompts. SD+\ours outperforms both SD~(\cfg) and SD+\dnp for all metrics except IS, under which it has slightly inferior performance compared to SD+\dnp. SDXL~(\cfg) also mirrors this behavior where SDXL+\ours outperforms on all metrics except IS. This shows that SDXL+\ours generates images that are both prompt-compliant and natural.

\Cref{tab:im} shows a similar comparison for the ImageNet dataset. For all datasets, SDXL+\ours outperforms SDXL~(\cfg) and SDXL+\dnp across all metrics.  To ensure that \ours improves correctness and quality without reducing the diversity of generated images, we also show FID on the ImageNet dataset in \Cref{tab:im}. Note that SDXL+\ours has a lower FID score (which implies high diversity) than SDXL~(\cfg) and SDXL+\dnp while maintaining higher prompt adherence.

While these metrics show that sampling with \ours improves the quality and prompt adherence of the images generated by all models, they do not convey an intuitive understanding of the significance of these gains. This is mostly because relatively small differences in the metrics can correspond to significant improvements in image synthesis. To provide better insight into the performance of the methods, we show win rate of each method in parenthesis \Cref{tab:pkdb}. SDXL+\ours wins $41-52\%$ of the time with the remaining split between SDXL~(\cfg) and SDXL+\dnp. These results show SDXL+\ours is preferred $2\times$ over SDXL~(\cfg) or SDXL+\dnp.

Human evaluations remain the gold standard for this assessment. For example, we have observed that adding \ours sampling nudges SDXL~(\cfg) toward more realistic images when the model is inclined to create paintings or cartoonish imagery. Generally, this makes images produced using \ours preferable to humans. \Cref{fig:humaeval} presents the results of the AMT evaluation, showing that humans prefer the image generated using SDXL+\ours over that synthesized by SDXL~(\cfg) for both quality and adherence. SDXL+\ours is preferred $46-61\%$ of the time, SDXL+\dnp is preferred $21-32\%$ of the time, and SDXL~(\cfg) is preferred $8-20\%$ of the time. Most \textit{`No clear winner'} cases correspond to situations where the prompt was too easy (all methods created similar or good enough images) or too hard (all methods failed to generate a correct image). This corroborates HPSv2's 2-to-1 win rate of SDXL+\ours over the baselines. When the prompt is too hard, increasing the value of $K$ in \ours allows corrections that no amount of guidance scale increase in \cfg can achieve. However, the results above do not capture this advantage since $K$ was kept fixed in these experiments. 

\subsection{Qualitative Results}
Qualitatively, we show the impact of adding \ours sampling to SDXL~(\cfg) for sample prompts all datasets in \Cref{fig:teaser}, where each image pair was synthesized with the same seed and guidance scale. These examples show the extent to which \ours produces better-quality images, illustrating how it is a general solution to many problems of current diffusion models. In particular, it shows that while SDXL~(\cfg) tends to generate paintings or cartoons for unnatural or complex prompts like ``a confused grizzly bear", SDXL+\ours is more inclined towards realistic or lifelike images. This is also evident in the prompts ``a cat and a dog" and ``a mouse and a pink bow". 

We observe that SDXL+\ours can add missing objects with their corresponding attributes, for example, the dog in ``a cat and a dog" and the glasses in ``a white bench and purple glasses". SDXL+\ours can also help with one of diffusion's most well-known weaknesses, anatomy, e.g., generating more anatomically correct hands, with no extra or missing fingers, as shown in the images for ``a girl diving into the pool" and ``a diamond ring on a girl's hand". Finally, \ours sampling allows SDXL~(\cfg) to better focus on spatial relations such as ``to the left of", text rendering, and numeracy.

Even for rare or imaginary prompts such as ``Burger with wheels on a racetrack," shown in \Cref{fig:more_result}, where SDXL shows extreme resistance towards generating this unreal scenario, maximizing the odds ratio using \ours forces compliance with the input prompt. For the prompt ``Dog and Santa. Black and white Christmas trees in the background", \ours not only ensures the presence of a dog and Santa Claus, but also enforces the complex requirement of black and white Christmas trees. 

\section{Ablations}

\begin{figure}[t]
\centering
\resizebox{\linewidth}{!}{
\begin{tabular}{c@{\hskip 0.1em}c@{\hskip 0.1em}c@{\hskip 0.1em}c}
\multicolumn{4}{c}{{\bf p} = {\tt $<$pixel art$>$ gray French bulldog}}\\
\includegraphics[width=.35\linewidth,valign=m]{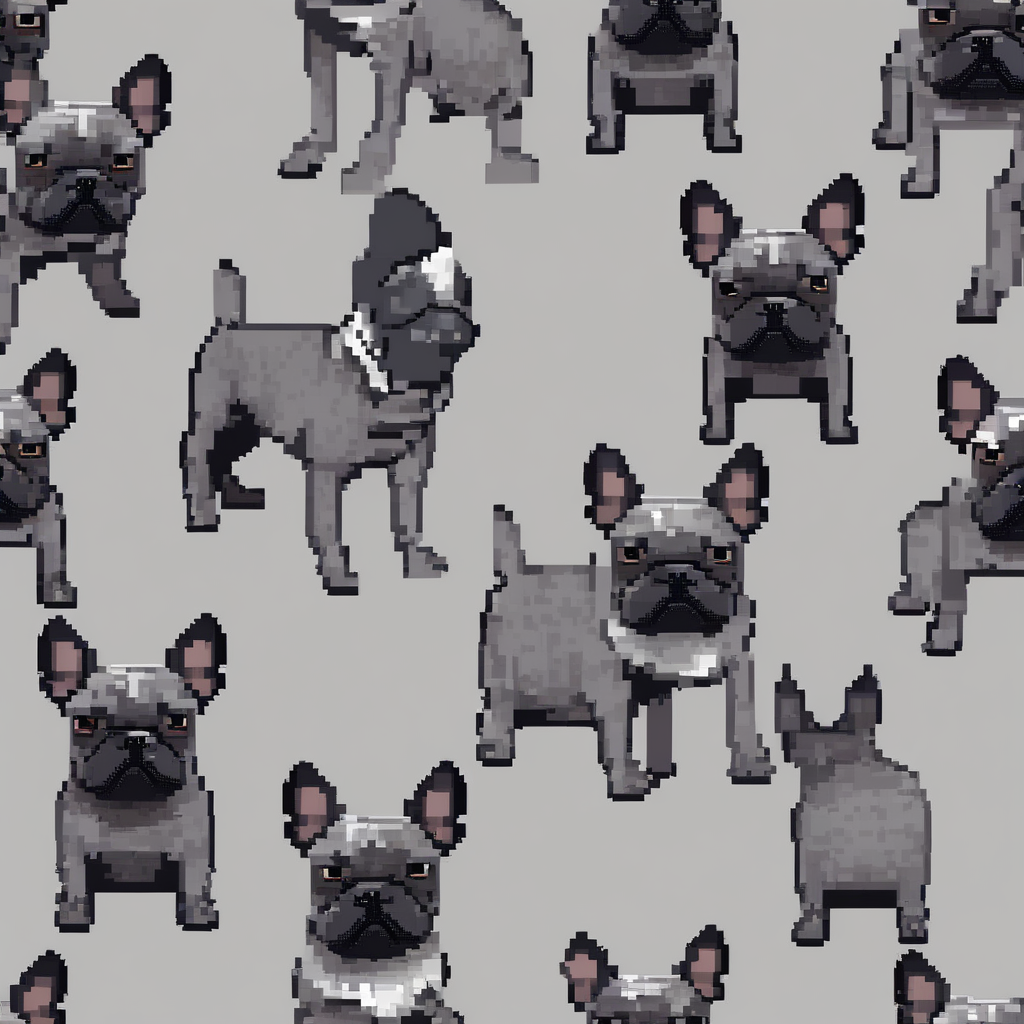} & 
\includegraphics[width=.35\linewidth,valign=m]{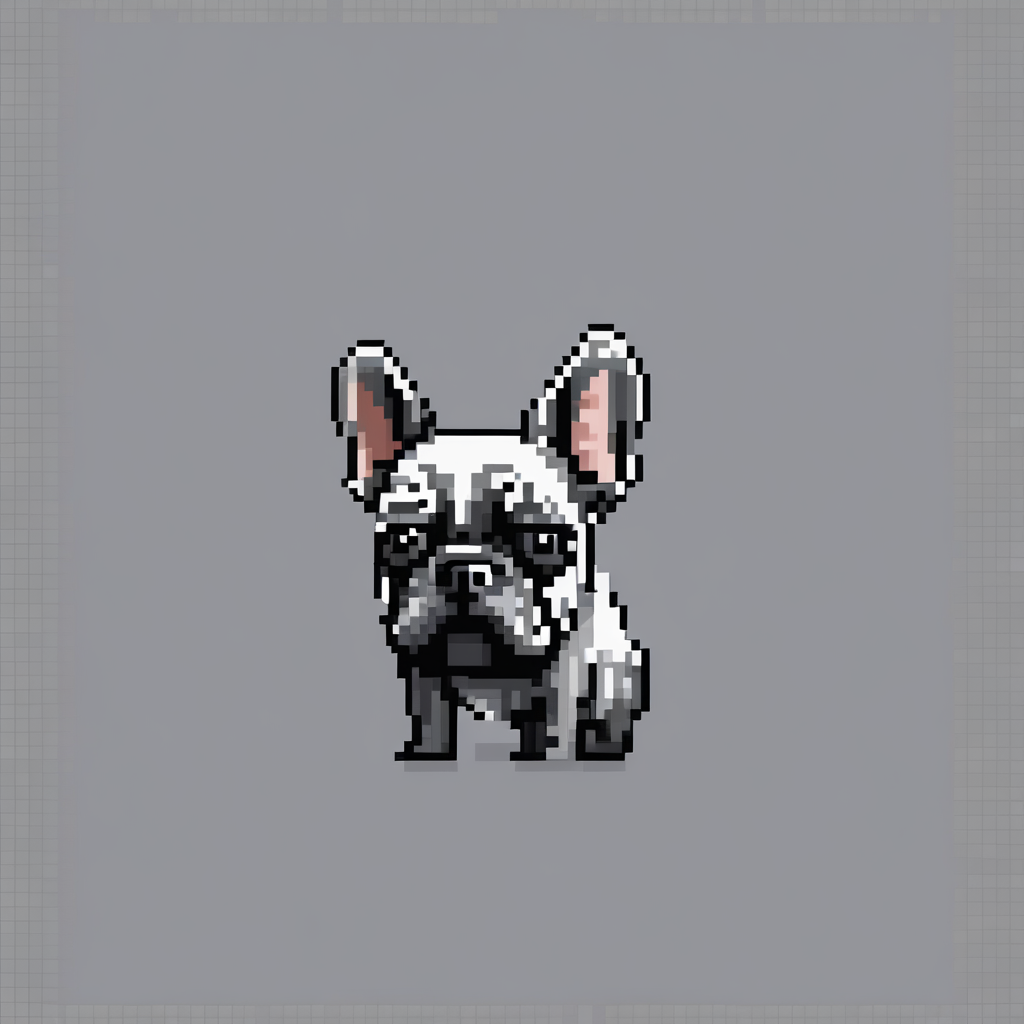} & 
\includegraphics[width=.35\linewidth,valign=m]{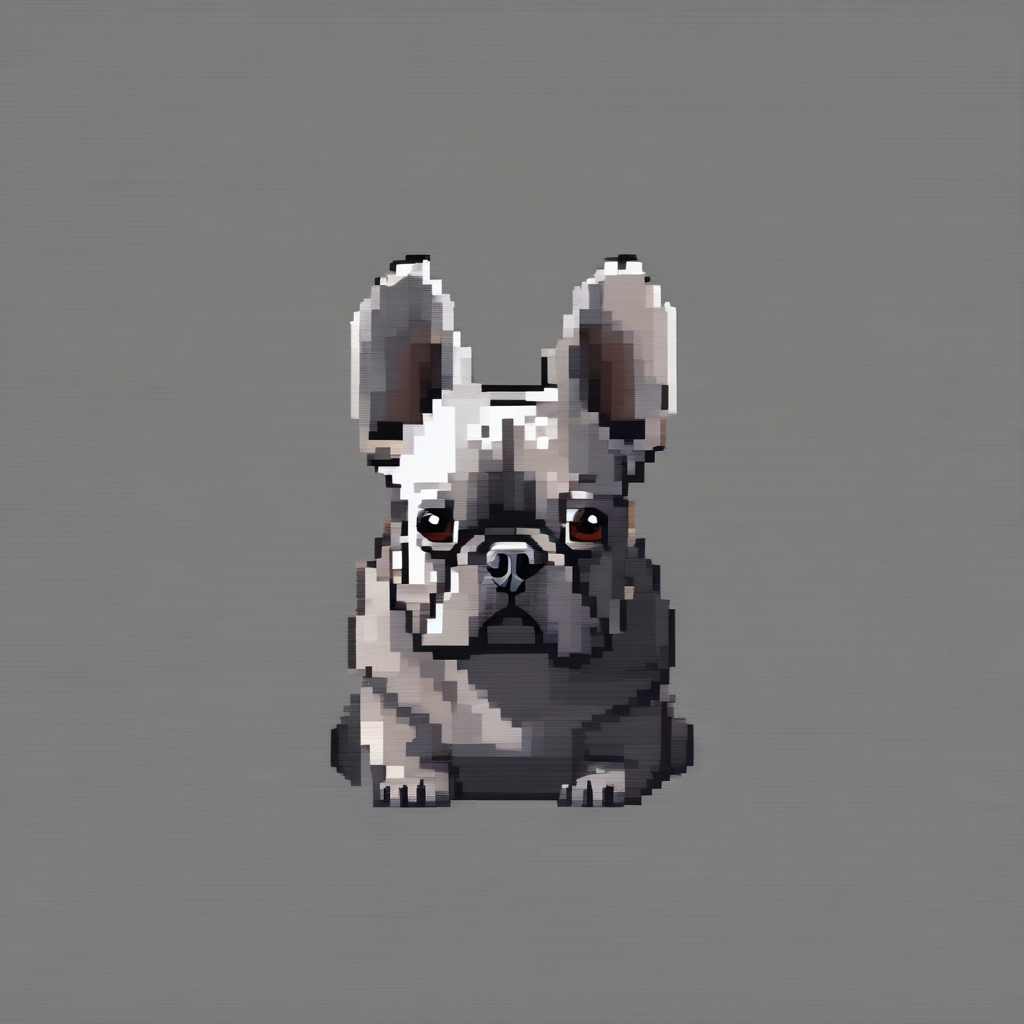} & 
\includegraphics[width=.35\linewidth,valign=m]{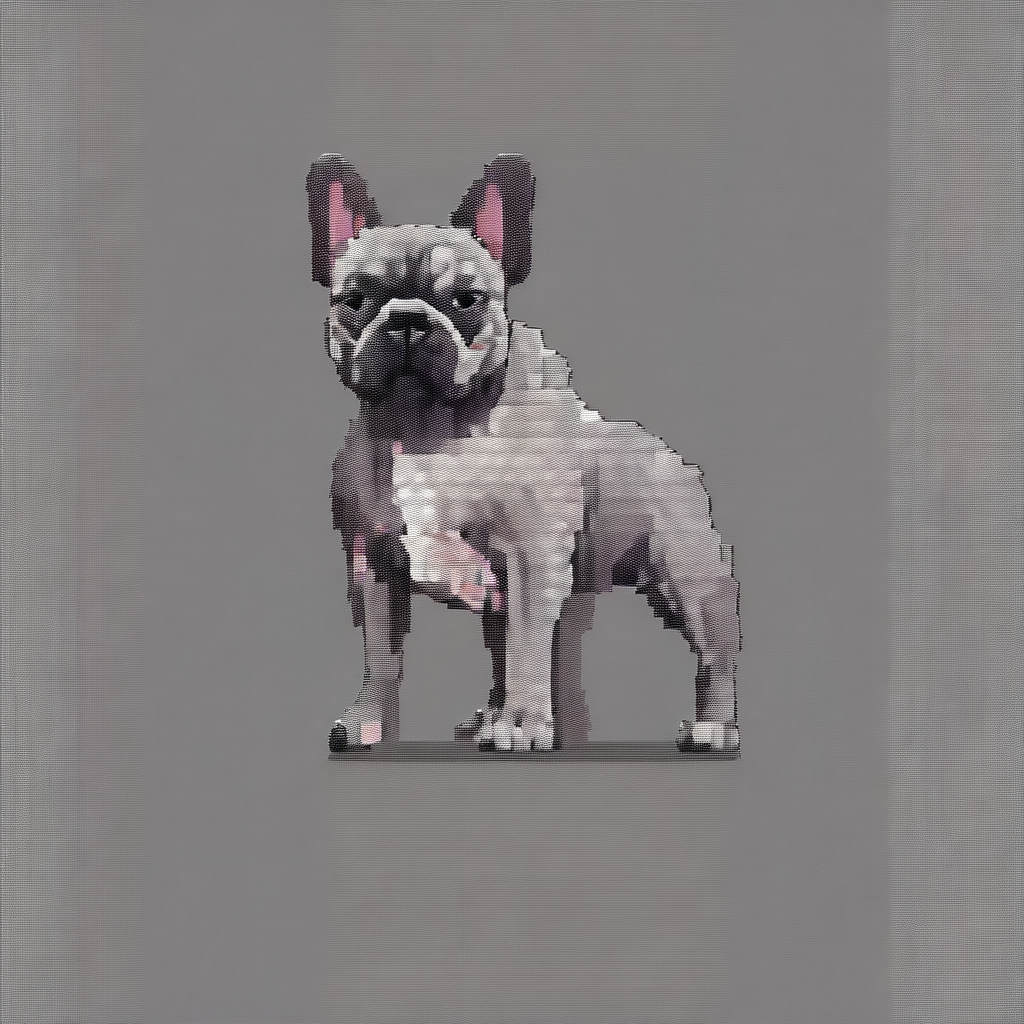} \\
SDXL~(\cfg) & \ours@K=5 & \ours@K=10 & \ours@K=15\\
\end{tabular}}
\vspace{-0.15in}
\caption{Effect of \dns hyperparameter $K$ on \ours.}
\label{fig:abls_n}
\end{figure}

\noindent\textbf{Number of DNS steps:}
The most important hyper-parameter of \ours is $K$, which determines the number of DNS steps performed at each iteration of the \np chain. While longer steps allow bigger corrections to the chain, there is a trade-off. When $K$ is too large, the mismatch between the DNS and \np chains produces a noise-smoothing effect that results in blurred images. In \Cref{fig:abls_k}, we show how FID and HPSv2 vary with run time for different $K$ using 
the SDXL~(\cfg) model and the ImageNet dataset~\cite{ILSVRC15}. Lower values of $K$ are equivalent to SDXL. As $K$ increases, both HPSv2 and FID improve. However, performance drops when $K$ is too high because the negative chain progresses too far, creating a large distribution mismatch and disrupting the Markov chain. This can be seen in the example of \Cref{fig:abls_n}, which shows the effect of the correction introduced by \ours on SDXL~(\cfg) as $K$ increases. While small values of $K$ improve prompt adherence, the image quality decreases beyond a certain threshold (for $K=15$ in this example). We empirically choose $K=5$ in our implementation and use it for all experiments in the paper. 

We note, however, that for individual prompts, other values of $K$ may achieve a better trade-off between prompt compliance and image quality. For simple prompts, SDXL suffices ($K=0$). For complex prompts, \ours enforces prompt adherence when SDXL fails even with the highest guidance scale. Overall, the throughput of SDXL+\ours can be controlled by limiting $K$ to as little as required for prompt compliance. More research is needed to automate the selection of $K$.

\begin{figure}[t]
\centering
\resizebox{\linewidth}{!}{
\begin{tabular}{c@{\hskip 0.1em}c@{\hskip 0.1em}c}
\multicolumn{3}{c}{{\bf p} = {\tt One car on the street}}\\
\includegraphics[width=.48\linewidth,valign=m]{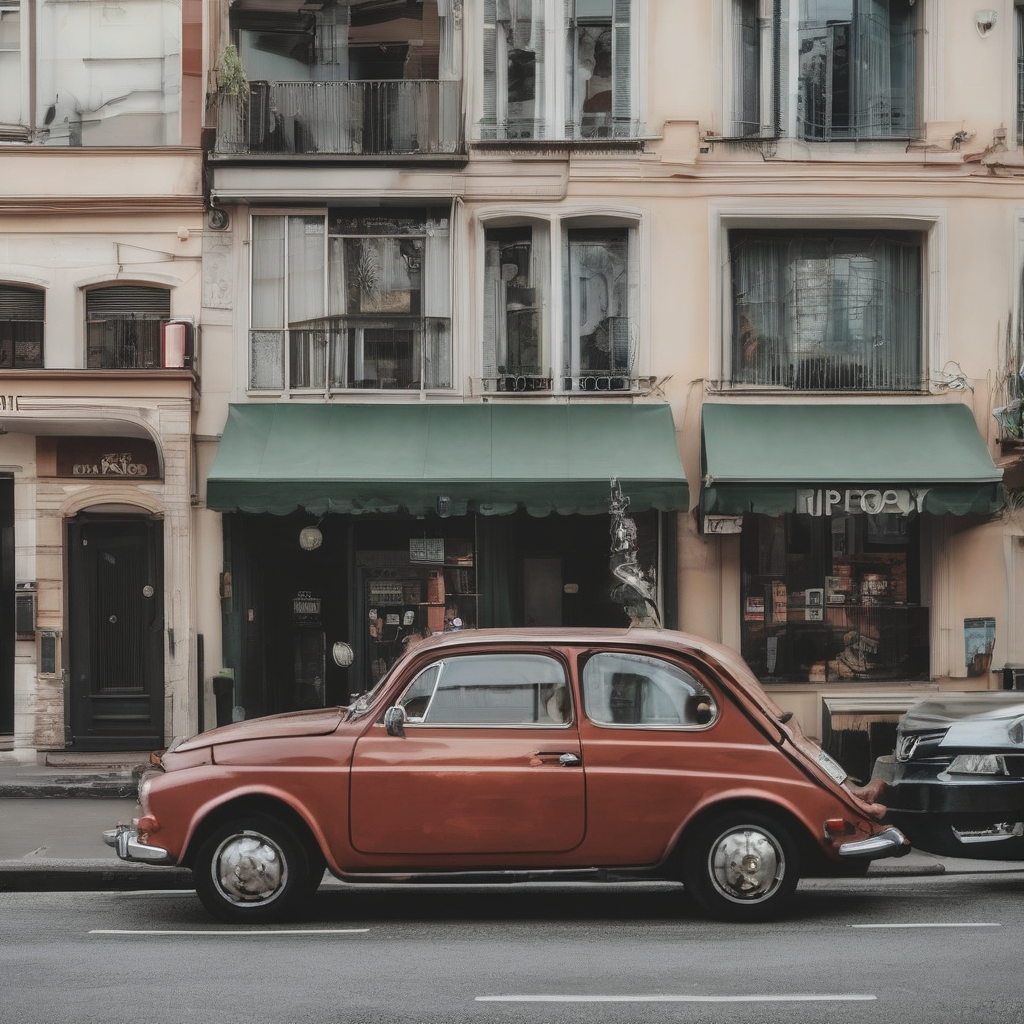} & 
\includegraphics[width=.48\linewidth,valign=m]{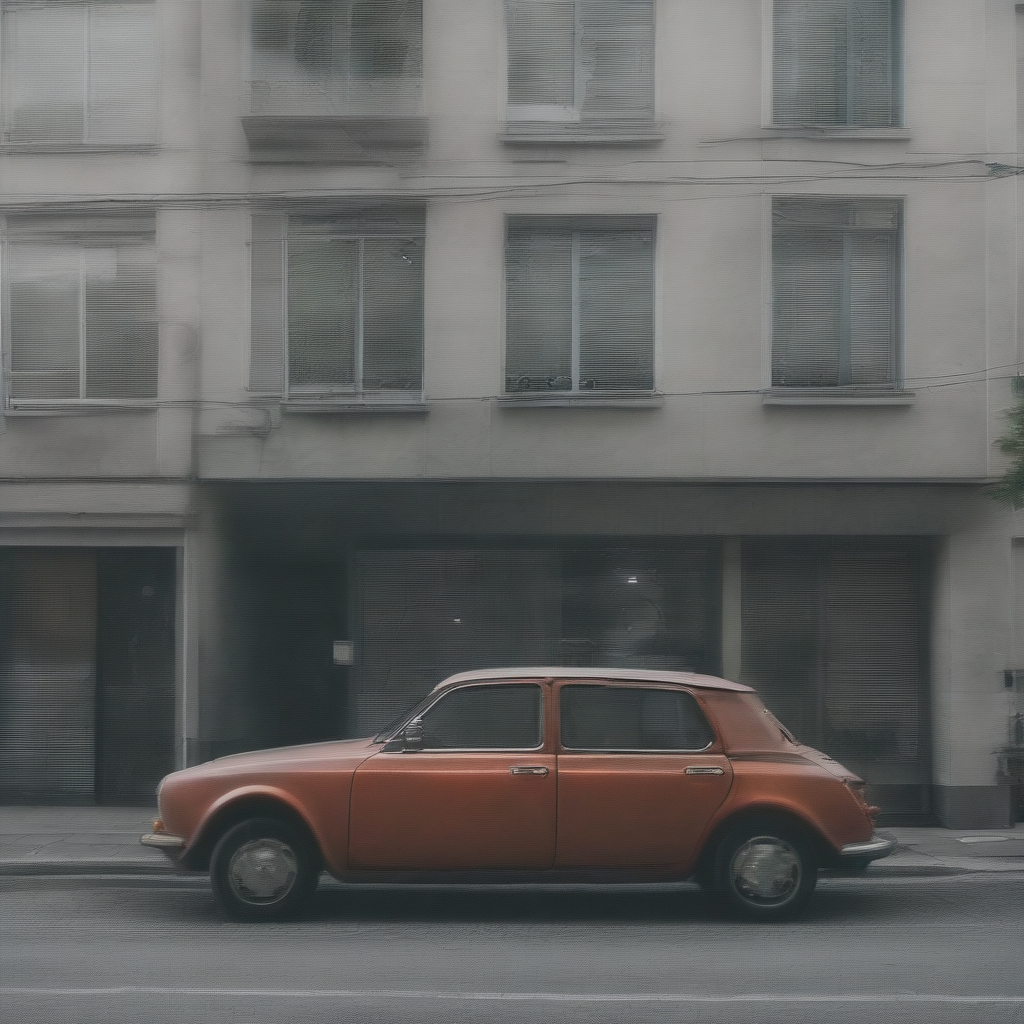} & 
\includegraphics[width=.48\linewidth,valign=m]{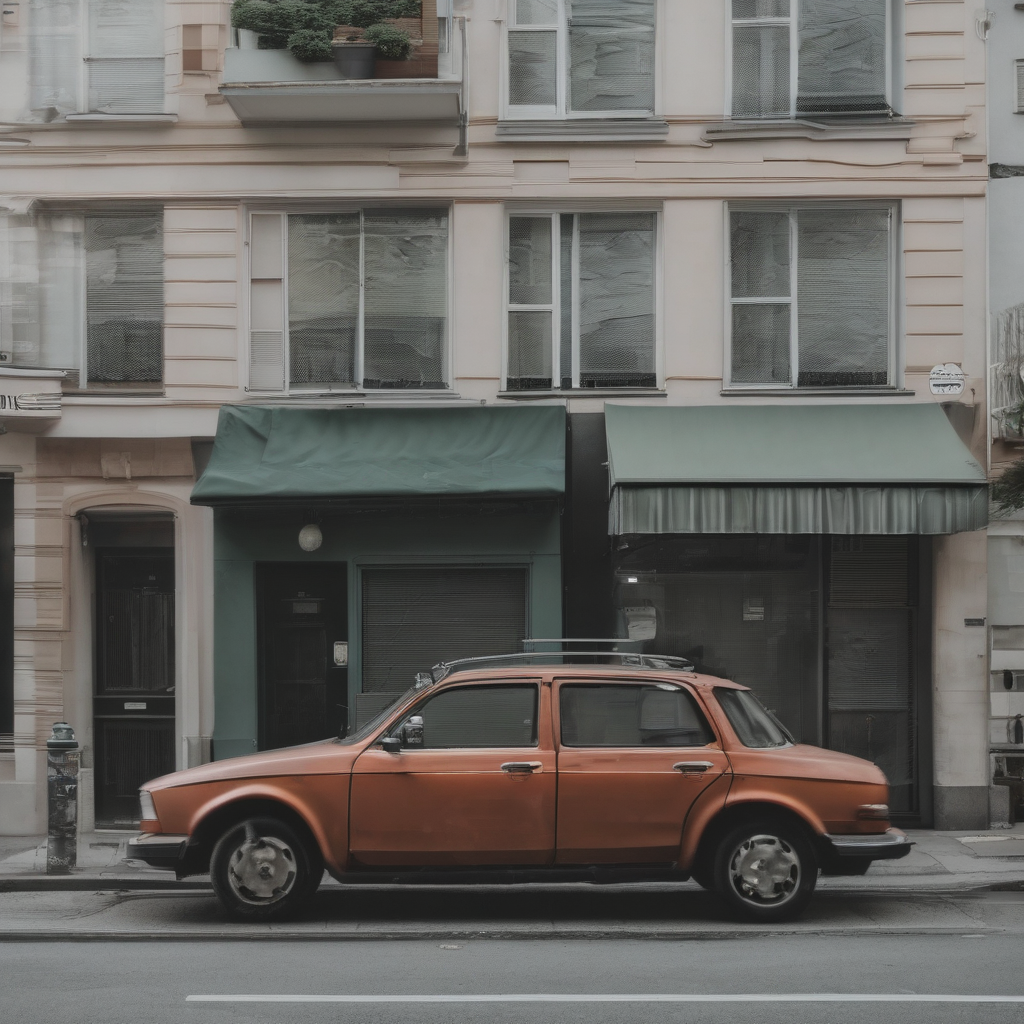} \\
SDXL~(\cfg) & \ours \textit{w/o normalization} & \ours\\
\end{tabular}}
\vspace{-0.15in}
\caption{Effect of normalization on \ours.}
\label{fig:abls_norm}
\end{figure}

\noindent\textbf{Effect of Normalization:}
\Cref{fig:abls_norm} shows the effect of the mean and variance normalization performed in \ours ($\it Normalize$ step of Algorithm~\ref{algo:ANS}). While \ours corrects the image and removes the front of a second car from the SDXL~(\cfg) image, it causes blurring when normalization is not used. This is because the DNS chain is $K$ steps ahead of the \np chain, which creates a mismatch in the noise statistics. Normalization corrects this and helps eliminate this blurring effect, as seen in \Cref{fig:abls_norm}.

\vspace{-0.05in}
\section{Related Work}
\vspace{-0.05in}
\noindent\textbf{Diffusion Models:} There is now a plethora of T2I models based on scalable model architectures~\cite{imagen, dalle, dalle2, ldm, sdxl} trained on large datasets~\cite{lion5b,coyo700m}. Many models leverage \cfg~\cite{cfg} to balance prompt compliance and diversity. This enables the synthesis of higher-quality images for more complex prompts. \ours will, in principle, benefit any model that leverages \cfg.

\noindent\textbf{Visual Conditioning \& Guidance:} Some works have introduced forms of conditioning other than text prompts, such as layouts~\cite{gligen, groundeddiff, layoutdiff}, example images~\cite{kumari2023multi, dreambooth, ragdiff}, sketches, or depth maps~\cite{controlnet}. While these conditions add structure and constraints to diffusion models (DMs), they are cumbersome or require skill. In \cite{fgdm} they address this by training a chain of models to generate visual conditions using adapters. However, these methods require additional resources and/or training. \ours requires no external resources and is intuitive since only a text prompt is needed.

\noindent\textbf{Text Conditioning:} Several approaches focus on text-only conditioning. Works like \cite{structdiff, compos} aim to correct errors induced by the DM text encoder by splitting the prompt into its constituent tokens or noun phrases, while, \cite{richtext} accepts Word document-like text formatting (e.g., font style, size, color, and footnotes) as input and applies the corresponding condition to each token based on the attention map. In \cite{ae, syngen}, they dynamically update latents by adjusting cross-attention between prompt tokens to enhance the presence and binding of attributes, respectively. These methods directly edit the latent at each step, which can significantly alter the Markov chain of the DM, frequently resulting in overly saturated and low-quality images. Recently, \cite{dnp} has shown that negative prompting with \dnp achieves better performance. Since \ours sampling uses \np sampling and updates latents based solely on DM-generated noise, it is much less prone to this problem.

\noindent\textbf{Negative Prompting:} The inference-time guidance introduced by \cite{compos} enables concept negation in diffusion models. Due to its effectiveness in filtering out unwanted elements from generated images, it has been widely embraced by most T2I models. However, crafting an effective negative prompt requires trial and error. To address this, \cite{dnp} proposed \dnp to learn a negative prompt optimized directly in diffusion space. Recently, \cite{qcomm} attempted to use corresponding text tokens as negatives by selectively removing them. Although effective, these techniques have limitations inherent to external captioning and latent manipulation by attention maps that do not affect \ours.
\vspace{-0.05in}
\section{Conclusion}
\vspace{-0.05in}
In this work, we hypothesize that the optimal negation for a diffusion model changes as the Markov chain proceeds. We then develop a sampling scheme, \ours, that uses adaptive negative sampling to correct the diffusion Markov chain without any external inputs or conditions. Our experiments show that updating the diffusion negative at each step generates images of higher quality and better prompt adherence. \ours requires no training or additional input and can be used with any model that supports classifier-free guidance.

\section{Acknowledgements}
This work was partially funded by NSF award IIS-2303153, NAIRR-240300, a gift from Qualcomm, and NVIDIA GPU donations. We also acknowledge and thank the use of
the Nautilus platform for the experiments discussed above.

{
    \small
    \bibliographystyle{ieeenat_fullname}
    \bibliography{main}
}

\end{document}